\documentclass[a4paper,11pt]{article}

\usepackage{authblk}
\usepackage[margin=4cm,columnsep=.8cm]{geometry}

\usepackage{algorithm2e}
\usepackage{amsthm}
\usepackage{amsmath}
\usepackage{amssymb}
\usepackage{balance}
\usepackage{graphicx}
\usepackage{soul}
\usepackage{url}
\usepackage{xcolor}
\usepackage{subcaption}
\DeclareMathOperator*{\argmax}{arg\,max}
\DeclareMathOperator*{\argmin}{arg\,min}

\newcommand{\myvec}{\boldsymbol}

\newcommand{\nonl}{\renewcommand{\nl}{\let\nl\oldnl}}

\newcommand{\power}{\mathsf{P}}
\newcommand{\load}{\mathsf{L}}
\newcommand{\demand}{\power_\text{dem}}
\newcommand{\turbine}{w}
\newcommand{\turbineset}{\mathcal{W}}
\newcommand{\nturbines}{{\left|\turbineset\right|}}
\newcommand{\subdepgraph}{G(\turbine)}
\newcommand{\setpoint}{s}
\newcommand{\atomicsetpoint}{s^\turbine}
\newcommand{\localsetpoint}{\myvec{\setpoint}^{\subdepgraph}}
\newcommand{\setpointset}{\mathcal{S}}
\newcommand{\localsetpointset}{\setpointset^{\subdepgraph}}
\newcommand{\methodacronym}{SPTS}
\newcommand{\powerfunc}[1]{\power^\turbine\left(#1\right)}
\newcommand{\powerfuncindex}[2]{\power^\turbine_{#2}\left(#1\right)}

\newcommand{\sptslearncurve}[4]{
	\begin{figure}[!h]
	\mbox{
	 \begin{subfigure}{0.24\textwidth}
	\includegraphics[scale=0.26,trim=0cm 0cm 0cm 0cm]{fig_appendix/#1_1.pdf}
	\caption{$n = 1$}
	\end{subfigure}
	 \begin{subfigure}{0.24\textwidth}
	\includegraphics[scale=0.26,trim=0cm 0cm 0cm 0cm]{fig_appendix/#1_2.pdf}
	\caption{$n = 2$}
	\end{subfigure}
	 \begin{subfigure}{0.24\textwidth}
	\includegraphics[scale=0.26,trim=0cm 0cm 0cm 0cm]{fig_appendix/#1_3.pdf}
	\caption{$n = 3$}
	\end{subfigure}
	 \begin{subfigure}{0.24\textwidth}
	\includegraphics[scale=0.26,trim=0cm 0cm 0cm 0cm]{fig_appendix/#1_4.pdf}
	\caption{$n = 4$}
	\end{subfigure}
	}
	\caption{#2 for $d = #3^\circ$ and $\power_\text{dem} = #4$ MW. The average trend and standard deviation are plotted.}
	\end{figure}
}

\newcommand{\sptsbest}[3]{
	\begin{figure}[!h]
	\mbox{
	 \begin{subfigure}{0.24\textwidth}
	\includegraphics[scale=0.26,trim=0cm 0cm 0cm 0cm]{fig_appendix/#1_1.pdf}
	\caption{$n = 1$}
	\end{subfigure}
	 \begin{subfigure}{0.24\textwidth}
	\includegraphics[scale=0.26,trim=0cm 0cm 0cm 0cm]{fig_appendix/#1_2.pdf}
	\caption{$n = 2$}
	\end{subfigure}
	 \begin{subfigure}{0.24\textwidth}
	\includegraphics[scale=0.26,trim=0cm 0cm 0cm 0cm]{fig_appendix/#1_3.pdf}
	\caption{$n = 3$}
	\end{subfigure}
	 \begin{subfigure}{0.24\textwidth}
	\includegraphics[scale=0.26,trim=0cm 0cm 0cm 0cm]{fig_appendix/#1_4.pdf}
	\caption{$n = 4$}
	\end{subfigure}
	}
	\caption{#2 for $d = #3^\circ$.}
	\end{figure}
}

\newcommand{\sptsfarmpower}[5]{
	\begin{figure}[!h]
	\mbox{
	 \begin{subfigure}{0.24\textwidth}
	\includegraphics[scale=0.2,trim=0cm 0cm 0cm 0cm]{fig_appendix/#1_1_#2.pdf}
	\caption{$n = 1$}
	\end{subfigure}
	 \begin{subfigure}{0.24\textwidth}
	\includegraphics[scale=0.2,trim=0cm 0cm 0cm 0cm]{fig_appendix/#1_2_#2.pdf}
	\caption{$n = 2$}
	\end{subfigure}
	 \begin{subfigure}{0.24\textwidth}
	\includegraphics[scale=0.2,trim=0cm 0cm 0cm 0cm]{fig_appendix/#1_3_#2.pdf}
	\caption{$n = 3$}
	\end{subfigure}
	 \begin{subfigure}{0.24\textwidth}
	\includegraphics[scale=0.2,trim=0cm 0cm 0cm 0cm]{fig_appendix/#1_4_#2.pdf}
	\caption{$n = 4$}
	\end{subfigure}
	}
	\caption{Average farm-wide power production of the best solution achieved by #3 for $d = #4^\circ$ and demand $\power_\text{dem} = #5$ MW.}
	\end{figure}
}

\newcommand{\BibTeX}{\rm B\kern-.05em{\sc i\kern-.025em b}\kern-.08em\TeX}

\begin{document}

\title{\textbf{Scalable Optimization for Wind Farm Control using Coordination Graphs}}

\author[1,2,*]{Timothy Verstraeten}
\author[2]{Pieter-Jan Daems}
\author[1]{Eugenio Bargiacchi}
\author[1,3]{Diederik M. Roijers}
\author[1,4]{Pieter J.K. Libin}
\author[2]{Jan Helsen}
\affil[1]{Vrije Universiteit Brussel, Artificial Intelligence Lab Brussels, Elsene, 1050, Belgium}
\affil[2]{Vrije Universiteit Brussel, Acoustics and Vibrations Research Group, Elsene, 1050, Belgium}
\affil[3]{HU University of Applied Sciences, Microsystems Technology, Utrecht, 3584CS, Netherlands}
\affil[4]{Data Science Institute, Hasselt University, Hasselt, 3500, Belgium}

\affil[*]{tiverstr@vub.be} 

\date{}
\maketitle

\begin{abstract}
Wind farms are a crucial driver toward the generation of ecological and renewable energy. Due to their rapid increase in capacity, contemporary wind farms need to adhere to strict constraints on power output to ensure stability of the electricity grid. Specifically, a wind farm controller is required to match the farm's power production with a power demand imposed by the grid operator. This is a non-trivial optimization problem, as complex dependencies exist between the wind turbines.
State-of-the-art wind farm control typically relies on physics-based heuristics that fail to capture the full load spectrum that defines a turbine's health status. When this is not taken into account, the long-term viability of the farm's turbines is put at risk.
Given the complex dependencies that determine a turbine's lifetime, learning a flexible and optimal control strategy requires a data-driven approach. However, as wind farms are large-scale multi-agent systems, optimizing control strategies over the full joint action space is intractable. We propose a new learning method for wind farm control that leverages the sparse wind farm structure to factorize the optimization problem. Using a Bayesian approach, based on multi-agent Thompson sampling, we explore the factored joint action space for configurations that match the demand, while considering the lifetime of turbines. We apply our method to a grid-like wind farm layout, and evaluate configurations using a state-of-the-art wind flow simulator. Our results are competitive with a physics-based heuristic approach in terms of demand error, while, contrary to the heuristic, our method prolongs the lifetime of high-risk turbines.
\end{abstract}


\section{Introduction}
As a scientific community, we acknowledge the environmental concerns associated with fossil fuel generation. Therefore, it is necessary to shift our energy generation to renewable energy sources \cite{climate_change_2018}. Wind energy is expected to play a major role, experiencing a yearly growth of about $10\%$ globally \cite{wwea_capacity_2019}.
By 2030, it is anticipated that at least $30\%$ of the European energy demand will be covered by wind energy \cite{wind_capacity_2030}.

The rapid increase in the supply of renewable energy poses challenges with respect to the stability of the electrical grid. In contrast to conventional power plants (e.g., gas, hydro and oil), the power output of wind farms ultimately depends on environmental conditions. Due to the increase in capacity, the integration of wind energy in the electricity grid needs to comply with strict grid code requirements \cite{sourkounis2013grid,ahmed2020grid}.

To ensure grid stability, wind farm controllers are developed to configure farm-wide power set-points, i.e., thresholds on the power production, in order to match the power demand \cite{tutorial_apc_2012}. This power demand is imposed by the transmission system operator, i.e., the entity responsible for balancing the energy supply and demand. The development of such controllers poses important challenges, as there exist complex non-linear dependencies between wind turbines. These dependencies originate from the wake effect \cite{wake_2012} in which upstream wind turbines reduce the available wind energy for downstream wind turbines. 
Additionally, when wind turbines are performing torque control, which regulates the power production by adapting the rotor speed, a higher power production typically results in increased loads on the mechanical components, which leads to a higher lifetime consumption \cite{iso_standard_design_2012}. Therefore, a careful balance between power and lifetime needs to be guaranteed \cite{tutorial_control_2017}.

The design of wind farm controllers is typically grounded in domain knowledge about patterns in the turbines' behaviors when the wake effect is present \cite{tutorial_control_2017,siniscalchi2019wind}. For example, one can recognize that upstream turbines, with respect to the dominant wind direction, typically observe higher fatigue loads (i.e., loads that induce weakening of the components) than downstream turbines \cite{jensen_fatigue_control_2016}. Therefore, lower set-points should be chosen for upstream turbines to reduce damage accumulation through fatigue loads, in case the available power over the farm is larger than the desired power.
While such heuristics simplify the computation of the optimal set-point allocation, they fail to capture the full complexity of the dynamically-changing multi-dimensional load spectrum (e.g., loads induced during storms). In order to develop advanced control strategies, it is necessary to consider the full load spectrum to reduce failures, which increases the reliability and sustainability of wind farms \cite{verstraeten2019fleetwide}.

In contrast to physics-based heuristics, data-driven wind farm controllers can learn control strategies without requiring in-depth knowledge about the non-linear dependencies that exist between the turbines that make up the wind farm. An example of such a data-driven structure is proposed in \cite{vandijk2016}, in which reinforcement learning techniques are used to search for the optimal rotor orientation to deflect wake away from downstream turbines. However, state-of-the-art data-driven wind farm controllers scale poorly to larger wind farms, as the number of possible configurations grows exponentially with respect to the number of wind turbines.

Therefore, we argue that a hybrid approach, combining both flexible data-driven methods and physics-based domain knowledge, is key to guarantee both optimality and scalability of wind farm controllers. We propose a new method that learns farm-wide control strategies in simulation while leveraging knowledge about wake patterns, performance statistics and load profiles. Specifically, we formalize the farm-wide dependencies caused by wake as a dependency graph and cluster wind turbines with similar load profiles together to factorize the wind farm. Using this factored representation, we propose a novel sampling method, called Set-Point Thompson Sampling (\methodacronym). This algorithm uses multi-agent Thompson sampling to evaluate promising control strategies using the factored representation of the wind farm, with the objective to match the power demand as well as possible, while minimizing stress on wind turbines with a low remaining useful life. Additionally, we use a Bayesian formalism, which allows the inclusion of available knowledge about the data in the form of prior belief distributions. Specifically, as the expected power production of a single turbine under no-wake conditions is readily available in the turbine's design specifications \cite{lydia2014comprehensive}, we construct a prior distribution for every set-point, centered around the associated expected power production. This guides the learning process toward sensible power productions while allowing for sufficient exploration to find the optimal set-point under wake conditions.

We start by positioning our research within related work and argue that controller optimization in large-scale wind farms is necessary in Section~\ref{sec:related_work}. Next, we formalize the set-point configuration problem and the objective in Section~\ref{sec:problem}. Then, we provide background on Thompson sampling in Section~\ref{sec:thompson_sampling}, after which we describe \methodacronym, an efficient method to explore possible set-point configurations, in Section~\ref{sec:spts}. We evaluate our method on realistic wind farm settings using an extensive set of parametrizations in Section~\ref{sec:experiments}. Finally, we discuss the results, highlighting the benefits and limitations of the method, and conclude with future work in Section~\ref{sec:discussion}.

\section{Related Work}
\label{sec:related_work}

Wind farm control strategies have mainly focused on power-load optimization (i.e., strategies that maximize power production and minimize fatigue load) or active power control (i.e., strategies that determine power set-points to meet the demand set by the electricity grid or an operator) \cite{tutorial_control_2017}.
In both cases, the wake effect is an important factor to consider when selecting power set-points. For power-load optimization, data-driven optimization approaches typically focus on reducing the wake effect, such as wake redirection control and axial induction control. Wake redirection control is concerned with finding a joint rotor orientation of the wind turbines in order to redirect wake from downstream turbines \cite{vandijk2016, wagenaar2012controlling}.
Axial induction control is an approach to reduce the wake effect, by lowering the power set-points of upstream turbines to reduce energy extraction, and thus maintain a steady wind speed behind the turbines \cite{soleimanzadeh2012optimization, gebraad2015maximum}.

In this work, we focus on active power control to match the wind farm's total power output to the power demand \cite{tutorial_apc_2012}. Many heuristic approaches based on physical knowledge about the turbines and environmental conditions exist \cite{tutorial_apc_2012,spudic2010hierarchical, siniscalchi2019wind,jensen_fatigue_control_2016}. For example, one can notice that due to the wake effect, a higher power production for upstream turbines results in lower wind speeds for downstream turbines. Therefore, the power demand can be reached using a heuristic approach, where the power contributions of downstream turbines are maximized while the power contributions of upstream turbines are minimized \cite{siniscalchi2019wind}.

We argue that, similar to the power-load optimization case, data-driven optimization approaches can complement existing physics-based knowledge to improve the flexibility of active power control. Such flexibility is important, as wind farm control decisions must consider the complex multi-dimensional load profiles to improve reliability \cite{verstraeten2019fleetwide}.
Nevertheless, it remains challenging to scale data-driven optimization methods to larger wind farms, where the optimal joint configuration exists in a high-dimensional solution space.

\section{Problem Statement}
\label{sec:problem}

When the transmission system operator imposes a power demand, the wind farm controller needs to configure each wind turbine to a power set-point such that the total actual power production matches the demand as closely as possible. These set-points need to be chosen without assigning high-load set-points to wind turbines with a low remaining lifetime. The lifetime of a turbine is dependent on many load factors that can lead to failure. However, the link between specific loading conditions and failure is not sufficiently understood \cite{load_causes_2016,load_failure_2017}. Therefore, in this work, we assume that the wind farm operator constructs a cost-function that heavily penalizes high loads on high-risk turbines based on expert knowledge (see Section~\ref{sec:challenges}).
We formalize the setting as a tuple $\langle \turbineset, G, \mathcal{R}, \setpointset, \langle\power, \load\rangle, \demand \rangle$, which can be regarded as an extension of a multi-agent multi-armed bandit \cite{stranders2012dcops,bargiacchi2018learning}, where
\begin{itemize}
\item $\turbineset$ is a set of wind turbines.
\item $G$ is a directed dependency graph that describes which subset of turbines influences a particular reference turbine. We refer to the dependencies of a wind turbine $\turbine \in \turbineset$ as its \emph{parents}, and denote the set that comprises turbine $\turbine$ and its parents as $\subdepgraph$.
\item $\mathcal{R}$ is a set of operational zones, or \emph{regimes}, within the wind farm, in which turbines observe similar loads under normal operating conditions. A fraction of the power demand will be allocated to each of these regimes.
\item $\setpointset = \setpointset^1 \times \dots \times \setpointset^{\nturbines}$ is the set of joint set-point configurations, which is the Cartesian product of the turbine-specific set-points $\atomicsetpoint \in \setpointset^\turbine$ for each turbine $\turbine \in \turbineset$. We denote $\localsetpointset$ as the set of local joint set-points for the set $\subdepgraph$.
\item $\power(\myvec{\setpoint})$ is a stochastic function providing the farm-wide power production when a joint set-point configuration, $\myvec{\setpoint} \in \setpointset$, is evaluated. The global power production can be decomposed into $\nturbines$ observable and independent local functions, i.e., $\power(\myvec{\setpoint}) = \sum\limits_{\turbine \in \turbineset} \powerfunc{\localsetpoint}$. The local function $\powerfunc{\localsetpoint}$ represents the power production achieved by wind turbine $\turbine$ and only depends on the local joint set-point $\localsetpoint$ of the subset of wind turbines in $\subdepgraph$.
\item $\load^\turbine\left(\localsetpoint\right)$ assigns a penalty for performing high-load actions by wind turbine $\turbine$. We assume that the wind farm operator heavily penalizes high-risk turbines (e.g., machines that have a low remaining life) based on available domain knowledge.
\item $\demand$ is the power demand imposed by the transmission system operator.
\end{itemize}

We aim to find the joint set-point configuration that matches the power demand as well as possible, while penalizing high-load actions on wind turbines with low remaining useful life:
\begin{equation}
\begin{split}
&\min_{\myvec{\setpoint}} \sum_{r \in \mathcal{R}} \left|f_r \demand - \sum_{w \in r} \power^\turbine\left(\localsetpoint\right)\right| + \sum_{\turbine \in \turbineset} \load^\turbine\left(\localsetpoint\right),
\end{split}
\label{eq:objective}
\end{equation}
where $f_r$ is a parameter that assigns a fraction of the demand to regime $r$.

\section{Thompson Sampling}
\label{sec:thompson_sampling}

Thompson sampling (TS) is a decision making strategy, in which the exploration process for the optimal action is guided by the user's prior beliefs about the unknown rewards of the actions.
Specifically, using a Bayesian formalism, it models the unknown parameters of the reward distribution for various pre-defined actions as prior belief distributions. The user should provide these distributions and quantify the existing uncertainty and information about these parameters, prior to learning.
Once a particular action has been evaluated and its reward has been measured, TS uses Bayes' rule to update the belief distributions, and thus maintains posteriors over the unknown parameters. Given a prior $Q_a$ for action $a$, and a history of observations $\mathcal{H}_{t-1}$, the user's beliefs about the expected reward $\mu(a)$ at time $t$ is modeled as:
\begin{equation}
\mu(a) \sim Q_a(\cdot\ |\ \mathcal{H}_{t-1}).
\end{equation}

TS is a probability matching mechanism \cite{lattimore_szepesvari_2020}, as it directly samples an action according to the probability that it is optimal at time step $t$. Specifically, for each action, it draws a sample from the associated posterior distribution, representing the expected reward of the action. Afterwards, it computes the action with the highest expected reward over all samples. Formally,
\begin{equation}
\begin{split}
\mu_t(a) &\sim Q_a(\cdot\ |\ \mathcal{H}_{t-1}), \forall a\\
a_t &= \argmax_{a} \mu_t(a).
\end{split}
\end{equation}
Finally, the chosen action is evaluated, and its measured reward (which may be stochastic) is added to the history of observations.

Note that the chosen action is independent and identically distributed with respect to the optimal action according to the user's beliefs \cite{lattimore_szepesvari_2020}, i.e.,
\begin{equation}
p(a_t\ |\ \mathcal{H}_{t-1}) = p(a_*\ |\ \mathcal{H}_{t-1}),
\end{equation}
where $a_*$ is the optimal action. Thus, TS balances exploration and exploitation by focusing on promising actions, while considering the user's uncertainty about the problem.
By allowing the user to configure prior belief about the problem, TS typically performs well in practical applications, where such information is often readily available \cite{chapelle2011empirical,libin2018bayesian,bfts2019}.

In multi-agent settings, the joint action space scales exponentially with the number of agents \cite{claus1998dynamics}. As TS does not leverage a factored representation of the joint action space, it becomes intractable to use for large multi-agent systems. Therefore, it is important to exploit the structure of the multi-agent setting whenever available, in order to decrease the complexity of the optimization problem \cite{kok2004scql,bargiacchi2018learning}.

\section{Set-Point Thompson Sampling}
\label{sec:spts}

To ensure scalability of the control optimization process toward large wind farms, we need a learning algorithm that can decompose and exploit the farm's topology. To this end, we propose a new wind farm control algorithm, Set-Point Thompson Sampling (\methodacronym), which constructs a dependency graph and regimes from wake fields and load profiles, and relies on multi-agent Thompson sampling (MATS) \cite{mats_journal_2020,mats_abstract_2020} to efficiently explore the joint set-point configuration space using the factored representation.

\subsection{Factorization}

To accurately decompose the problem, we rely on two aspects.
Firstly, wind turbines depend on each other due to the wake effect. The decisions made by upstream turbines affect downstream turbines. Therefore, local coordination between a reference turbine and its affected neighbors is necessary to guarantee optimality of the solution.
Secondly, due to the wake effect, upstream wind turbines produce more power than downstream turbines. As power and torque loading are highly correlated, wind turbines with similar power productions overall have similar fatigue load profiles \cite{alvarez2018improved,verstraeten2019fleetwide}. 
Since turbines with a lower observed fatigue load should be responsible for the majority of the demand, a group of similar turbines can be assigned a fraction of the demand inversely proportional to their observed loads.

\begin{figure}[!htb]
\centering
\includegraphics[scale=0.35,trim=2.5cm 0cm 0cm 0cm]{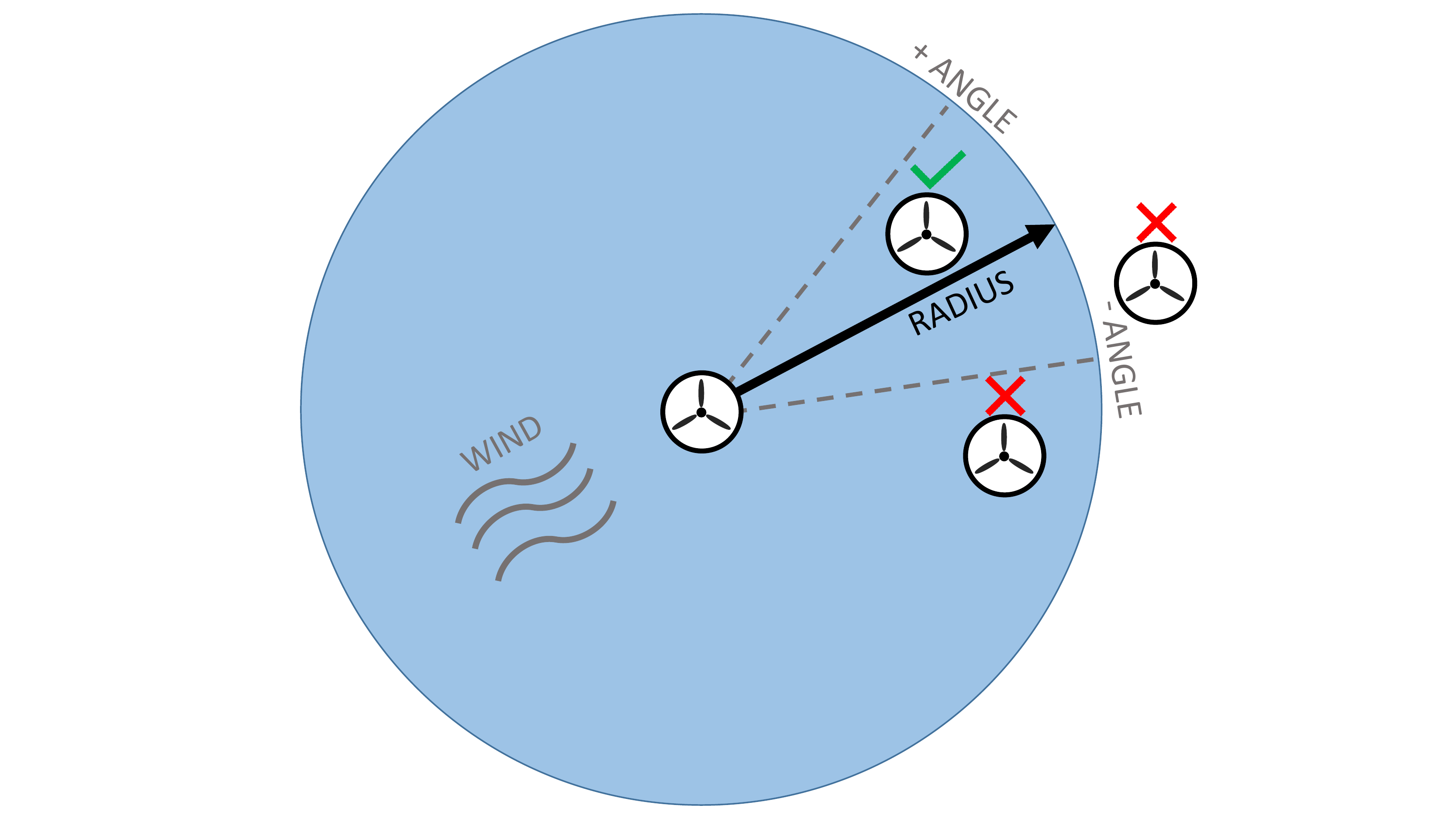}
\caption{An example with one upstream reference turbine and three downstream turbines. Only one downstream turbine is considered to be dependent on the reference turbine (green check mark), as it is both within the specified radius and within the specified angle with respect to the incoming wind vector from the reference turbine.}
\label{fig:example_graph_inclusion}
\end{figure}

To construct the dependency graph $G$, we analyze the wake field generated at a given wind direction. Using a geometric approach, we derive, for a particular reference turbine, which downstream turbines are in its wake. Specifically, we consider a downstream turbine $\turbine$ to be dependent on an upstream reference turbine $\turbine_\text{ref}$ if and only if $\turbine$ is geographically positioned within a specified radius from $\turbine_\text{ref}$ and is within a specified angle from the incoming wind vector. Naturally, a larger angle and radius would lead to a more dense dependency graph. For our purpose, we found that downstream turbines located at an angle of $22.5^\circ$ and a radius of $1$ km sufficiently lie outside of the wake effect induced by the reference turbine, according to the simple Jensen wake model \cite{katic1986simple}. A visualisation of the geometric approach is shown in Figure~\ref{fig:example_graph_inclusion}. Examples of a wake field and corresponding dependency graph are shown in Figure~\ref{fig:setup}.

To construct the operational regimes $\mathcal{R}$, we group wind turbines based on their observed fatigue loads. As mentioned before, turbines with similar overall power productions have similar load profiles. Since the majority of turbine operations are performed in steady states under the dominant wind direction, we cluster the turbines based on their power productions under active wake conditions. Specifically, we cluster the turbines using the state-of-the-art approach described in \cite{verstraeten2019fleetwide}, which uses a Bayesian Gaussian mixture model. These models describe the data using a finite set of multivariate Gaussian distributions. Afterwards, each turbine is assigned to the regime (i.e., Gaussian) it most likely belongs to.

Once the regimes have been created, we assign a fraction $f_r$ of the demand to each regime $r$. This fraction should be proportional to the remaining useful life of the regime. 
We construct the load revolution distribution \cite{iso_standard_design_2012}, i.e., the number of rotations performed by the turbines within a regime operating at a particular torque level, for each operational regime, based on real wind farm data of 24 turbines. We define the remaining lifetime in terms of the accumulated damage, which is the number of rotations operated under high-load conditions. We consider an operation to be high-load when the expected (main shaft) torque exceeds a particular threshold. Formally, the torque of turbine $w$ at time $t$ is defined as
\begin{equation}
\tau^\turbine_t = \frac{\power^\turbine_t}{\omega^\turbine_t},
\end{equation}
where $\power^\turbine_t$ and $\omega^\turbine_t$ are, respectively, the produced power (in W) and angular rotor speed (in rad/s) of turbine $w$ at time $t$. We assume that damage occurs when the torque exceeds the one achieved at 65\% of the turbine's maximum power level \cite{alvarez2018improved}.
Thus, we can derive the number of rotations performed at a torque higher than this threshold from the load revolution distribution of each turbine and aggregate the results per regime. After normalization of the remaining rotations over the entire wind farm, the fraction $f_r$ is set to the inverse of the sum over all turbines within regime $r$. An example of regimes, associated with the normalized number of life-consuming rotations per turbine, is shown in Figure~\ref{fig:setup_regimes}.

\subsection{Exploration}

Consider the problem described in Section~\ref{sec:problem}. The expected power productions $\powerfunc{\localsetpoint}$ for each possible local set-point configuration $\localsetpoint$ are unknown. Similar to MATS \cite{mats_journal_2020}, \methodacronym\ uses a Bayesian formalism, which means users can exert their beliefs over $\powerfunc{\localsetpoint}$ in the form of a prior.
If wind turbines were not affected by wake, the incoming wind speed can be used to predict the expected power production of the wind turbine, which is provided with the turbine design specifications \cite{lydia2014comprehensive}. Therefore, for a given set-point, wind speed and turbine, we model the achieved power using a Gaussian prior, where the mean is the expected power production given the set-point under no wake conditions, and the standard deviation $\sigma$ represents prior uncertainty about the achieved power.
\begin{equation}
\begin{split}
\powerfunc{\myvec{s}} &\sim \mathcal{N}(\cdot\ |\ \mu^\turbine_{\atomicsetpoint}, \sigma),\\
\mu^\turbine_{\atomicsetpoint} &= \min(\atomicsetpoint, \power^\turbine_\text{av}),
\end{split}
\label{eq:prior}
\end{equation}
where the mean is the minimum between the set-point $\atomicsetpoint$ and the available power $\power^\turbine_\text{av}$. The standard deviation $\sigma$ balances exploration of alternative power production outcomes (high $\sigma$), and exploitation of the provided domain knowledge (low $\sigma$).

At each time step $t$, \methodacronym\ draws a sample $\powerfuncindex{\localsetpoint}{t}$ from the posterior for each wind turbine and possible local set-point configuration, given the history, $\mathcal{H}_{t-1}$, consisting of previously evaluated set-points and associated power productions:
\begin{equation}
\begin{split}
&\powerfuncindex{\localsetpoint}{t} \sim \mathcal{N}(\cdot\ |\ \mu^\turbine_{\atomicsetpoint}, \sigma, \localsetpoint, \mathcal{H}_{t-1}),\text{ with}\\
&\mathcal{H}_{t-1} = \bigcup\limits^{t-1}_{i=1} \bigcup\limits_{\turbine \in \turbineset} \left\{\left\langle \localsetpoint_i, \powerfunc{\localsetpoint_i} \right\rangle\right\}.
\end{split}
\end{equation}

Note that during this step, \methodacronym\ samples directly the posterior over the unknown local means, which implies that the sample $\powerfuncindex{\localsetpoint}{t}$ and the unknown mean $\powerfunc{\localsetpoint}$ are independent and identically distributed at time step $t$, given history $\mathcal{H}_{t-1}$.

\methodacronym\ takes the set-point that minimizes the objective function (see Equation~\ref{eq:objective}).
In traditional Thompson sampling (TS) \cite{thompson1933likelihood}, the optimal solution is found by maximizing over the full joint action space. However, this is intractable for larger multi-agent settings, as the joint action space scales exponentially with the number of agents. For example, a moderately-sized wind farm comprised of 20 wind turbines, where each turbine can choose from 3 possible set-points, would have $3^{20}$ (approximately 3.5 billion) possible configurations.
Fortunately, due to the structure of wind farms, the optimal set-point configuration exists in the sparse factored representation of the joint action space. Therefore, the minimization problem defined in Equation~\ref{eq:objective} can be solved exactly and effectively using variable elimination \cite{guestrin2002multiagent} or linear programming \cite{coin_2003}.

Finally, the joint set-point configuration that minimizes the objective function, $\myvec{\setpoint}_t$, is executed in simulation and the associated power productions $\powerfuncindex{\localsetpoint_t}{t}$ will be recorded for each wind turbine $\turbine$. \methodacronym\ is formally described in Algorithm~\ref{algo:spts}.

\RestyleAlgo{boxruled}
\LinesNumbered
\IncMargin{0.1cm}
\begin{algorithm}[!b]
\caption{\methodacronym}
\label{algo:spts}
$G, \mathcal{R} \leftarrow$ Construct dependency graph and regimes\\
$\mathcal{H}_0 \leftarrow \{\}$\\
\For{$t \in [1..T]$}{
	\emph{Sample expected performance for every possible local set-point configuration.}\\
	\For{$w \in \mathcal{W}, \myvec{\setpoint} \in \localsetpointset$}{
	$\powerfunc{\myvec{s}} \sim \mathcal{N}\left(\ \cdot\ \middle|\ \mu^\turbine_{\atomicsetpoint}, \sigma, \mathcal{H}_{t-1}\right)$
	}\vspace{0.2cm}
	\emph{Select best joint configuration.}\\
	$\myvec{\setpoint}_t \leftarrow \argmin_{\myvec{\setpoint}} \sum\limits_{r \in \mathcal{R}} \left|f_r \demand - \sum_{w \in r} \powerfunc{\localsetpoint}\right|$\\
	\qquad \qquad \qquad $+ \sum\limits_{\turbine \in \turbineset} \load^\turbine\left(\localsetpoint\right)$\\\vspace{0.2cm}
	\emph{Simulate chosen set-point configuration.}\\
	$\left\langle \powerfuncindex{\localsetpoint_t}{t} \right\rangle_{\turbine \in \turbineset} \leftarrow $ Simulate configuration $\myvec{\setpoint}_t$\\\vspace{0.2cm}
	\emph{Update belief distributions using observed performance.}\\
	$\mathcal{H}_t \leftarrow \mathcal{H}_{t-1} \cup \left\{\left\langle \localsetpoint_t, \powerfuncindex{\localsetpoint_t}{t} \right\rangle_{\turbine \in \turbineset}\right\}$\\
}
\end{algorithm}

\section{Experiments}
\label{sec:experiments}

\begin{figure*}[!htb]
    \centering
    
    \begin{subfigure}{.49\textwidth}
    \centering
    \includegraphics[scale=0.4,trim=0.8cm 0cm 0cm 0cm,clip]{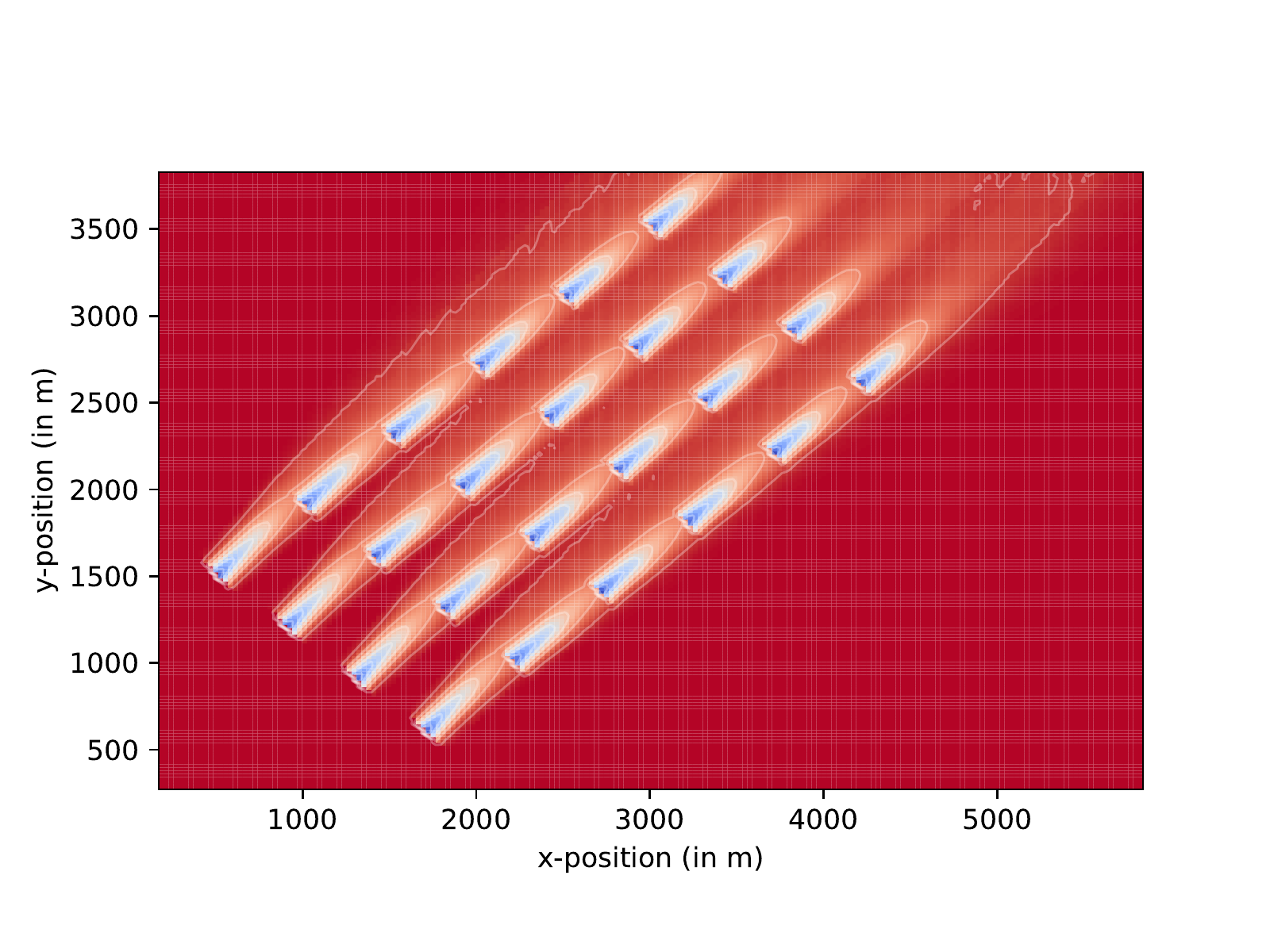}
    \caption{Wake field -- $0^\circ$}
    \end{subfigure}
    \begin{subfigure}{.49\textwidth}
    \centering
    \includegraphics[scale=0.4,trim=0.5cm 0cm 0cm 0cm,clip]{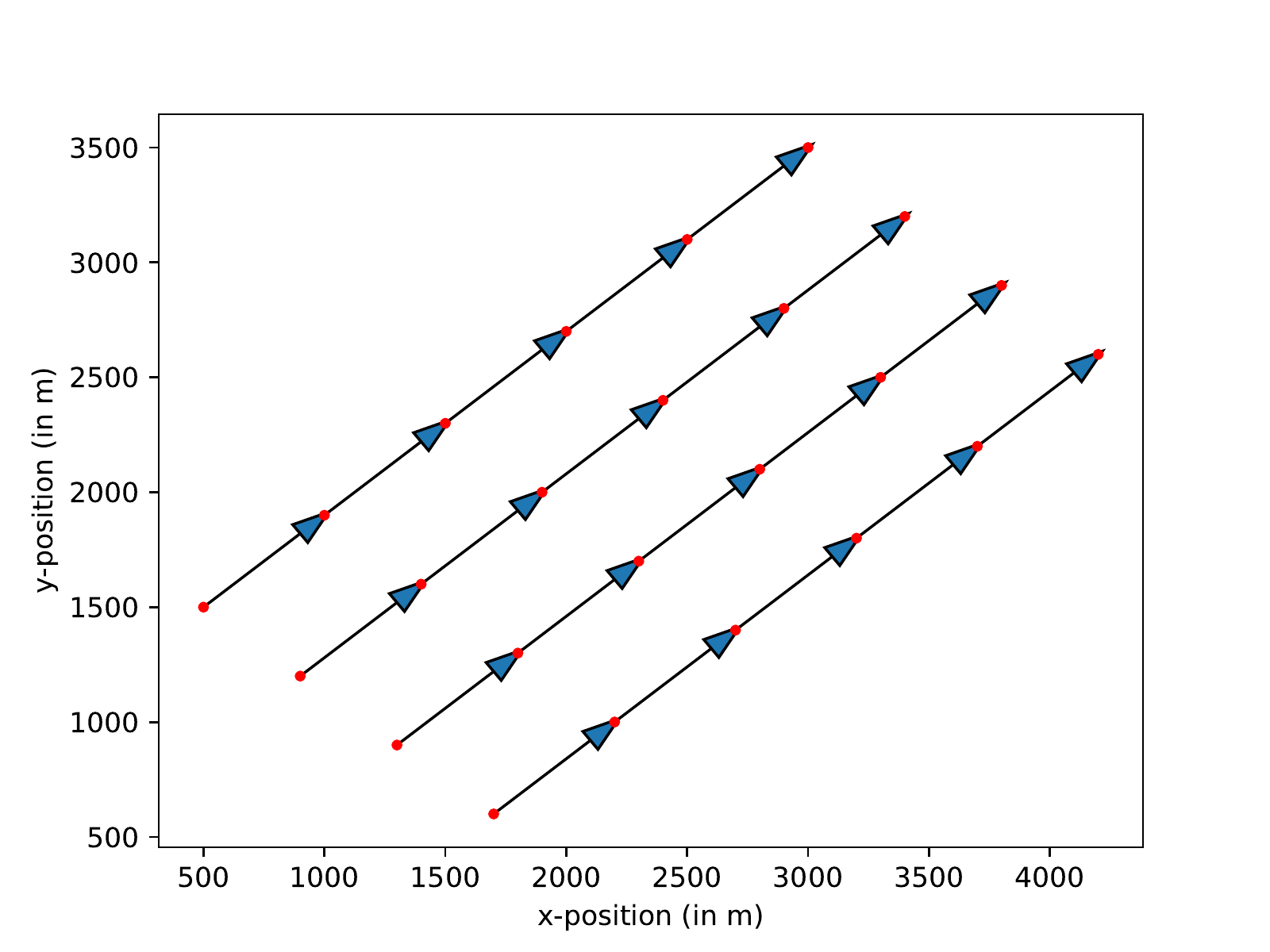}
    \caption{Dependency graph -- $0^\circ$}
    \end{subfigure}\\
    \begin{subfigure}{.49\textwidth}
    \centering
    \includegraphics[scale=0.4,trim=0.8cm 0cm 0cm 0cm,clip]{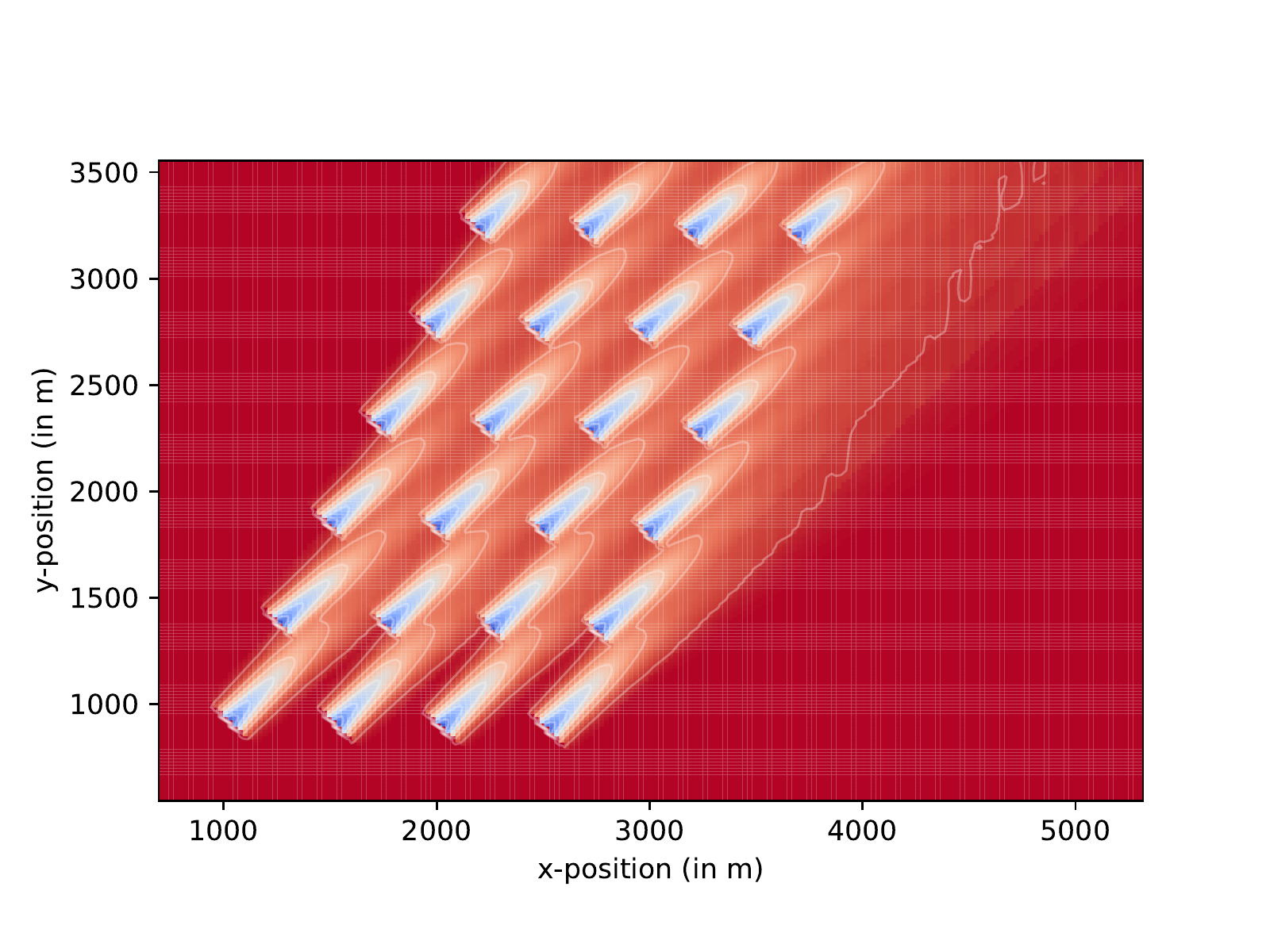}
    \caption{Wake field -- $30^\circ$}
    \end{subfigure}
    \begin{subfigure}{.49\textwidth}
    \centering
    \includegraphics[scale=0.4,trim=0.5cm 0cm 0cm 0cm,clip]{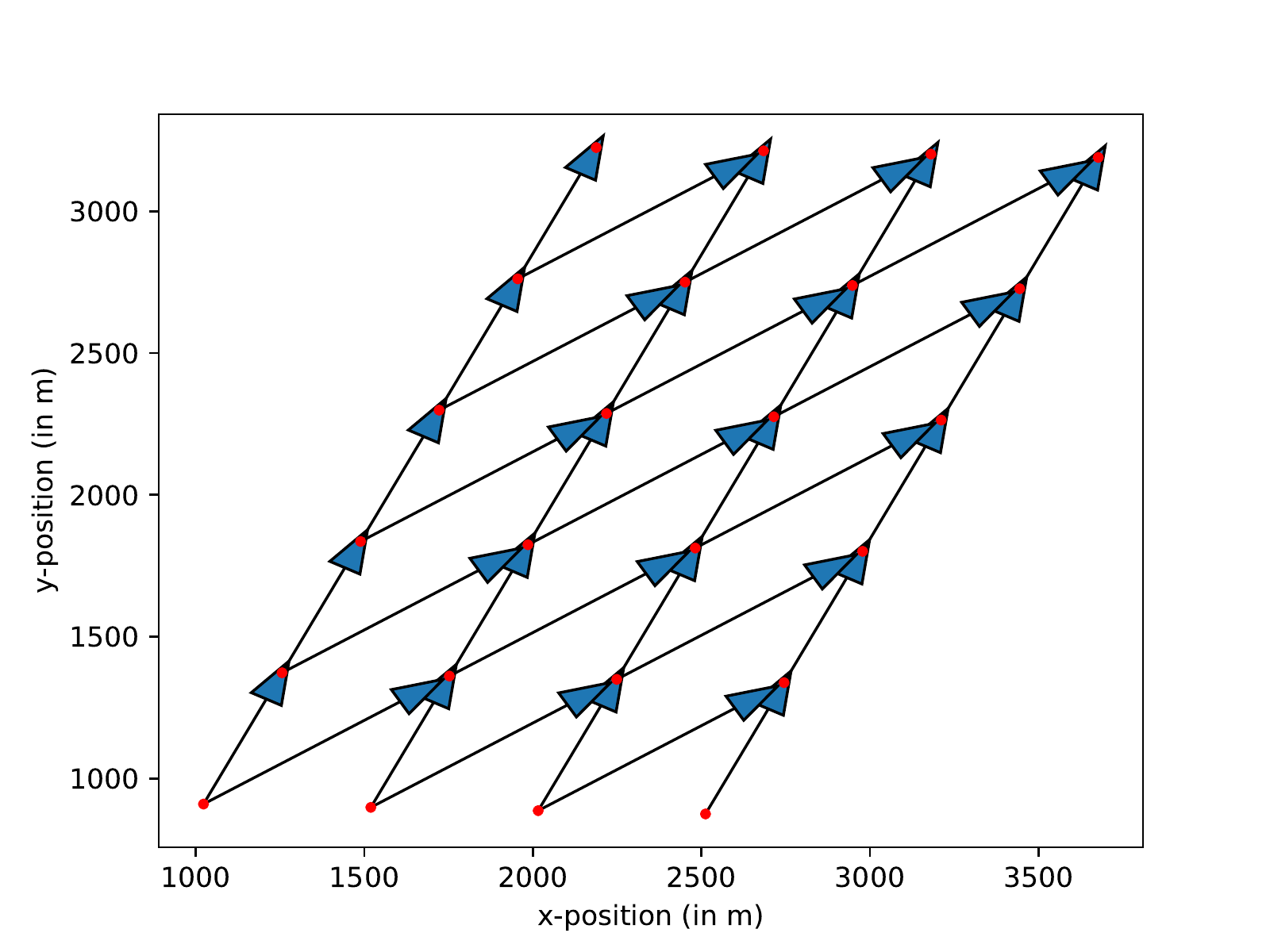}
    \caption{Dependency graph -- $30^\circ$}
    \end{subfigure}

  \caption{Wind farm -- Based on the wake field generated for a particular wind direction, wind speed of $11$ m/s, and maximal power set-points, we create a dependency graph. We show the results for wind directions of $0^\circ$ (a+b) and $30^\circ$ (c+d). For a wind direction of $30^\circ$, the wind farm is rotated by $-30^\circ$, such that the global wind vector always starts at $(0, 0)$.}
  \label{fig:setup}
\end{figure*}

As the majority of offshore wind farms have a symmetric grid-like topology \cite{tao2020topology}, we conduct our experiments in a wind farm that has the shape of a parallelogram. Grid-like layouts are often beneficial toward the planning and construction of the farms. However, such layouts cause wake due to the proximity of the turbines, reducing the overall power production of the wind farm \cite{tao2020topology}. We place 24 turbines in a 4-by-6 grid, 500 m apart along the x-axis and 400 m apart along the y-axis, as shown in Figure~\ref{fig:setup}.

\begin{figure}[!htb]
    \centering
    \includegraphics[scale=0.45]{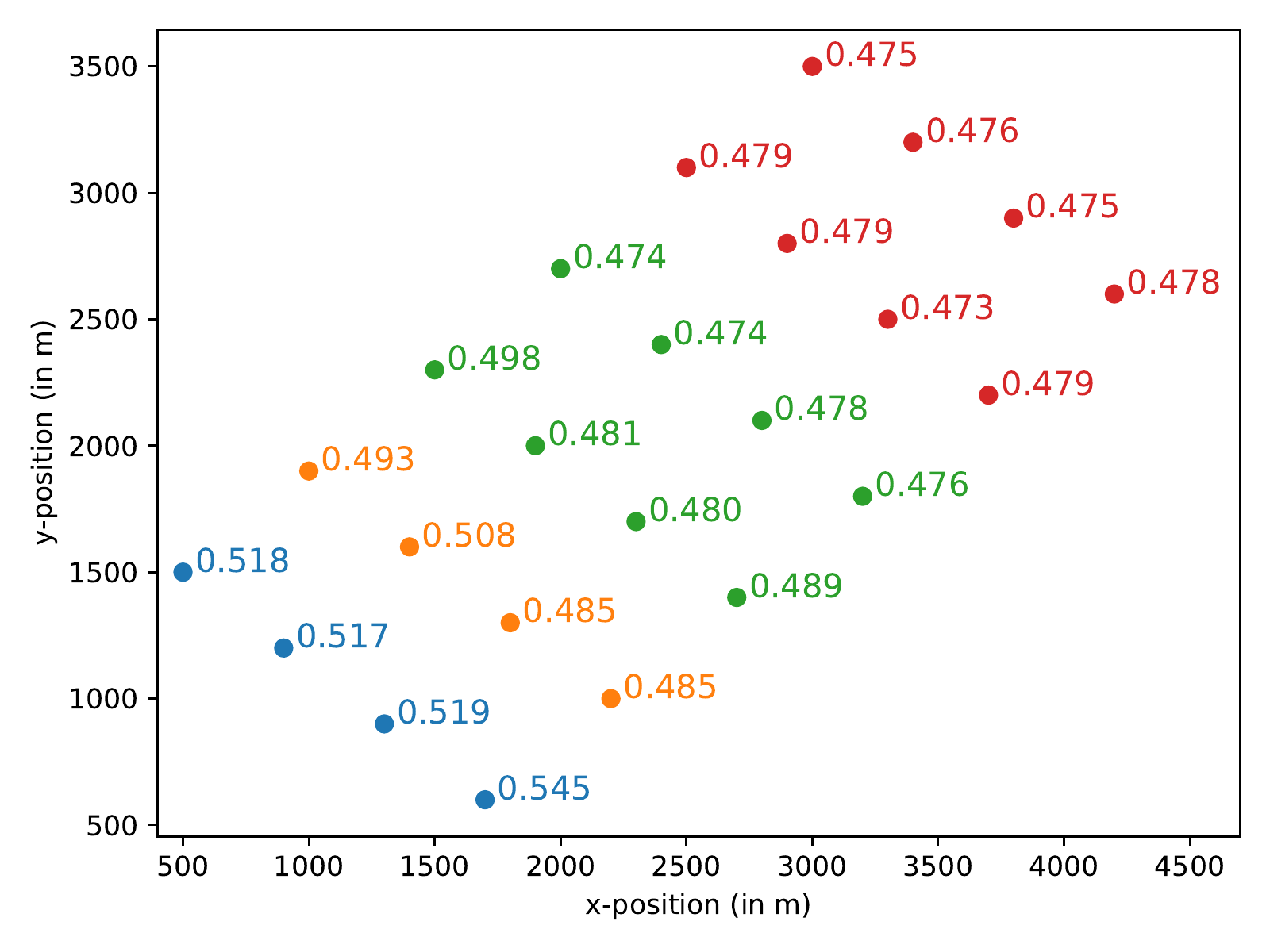}
  \caption{Operational regimes in the wind farm, obtained under the dominant wind vector coming from the origin $(0, 0)$. The turbines are annotated with their observed normalized damage-inducing load.}
  \label{fig:setup_regimes}
\end{figure}

We investigate our method under the same global wind conditions used by \cite{siniscalchi2019wind}, in which we assume that $0^\circ$ is the dominant wind direction and orthogonal to the wind farm grid. Specifically, we investigate the incoming wind vectors at $0^\circ$ and $30^\circ$, with a speed of $11$ m/s. Under these conditions, the wake effect is strong. The wake fields and dependency graphs for both wind vectors are shown in Figure~\ref{fig:setup}.

We use FLORIS \cite{floris_2019}, a state-of-the-art simulator (constructed by the National Renewable Energy Lab in the USA) to simulate farm and wake conditions, and the LW-8MW reference turbine \cite{desmond2016description} to model the turbines' dynamics. These models accurately represent the scale of, and conditions at, contemporary offshore wind farms. From real wind farm data, we compute damage-inducing fatigue loads for every turbine, according to \cite{alvarez2018improved}, and derive the fractions $f_r$ as discussed in Section~\ref{sec:spts}.
The normalized damage-inducing fatigue loads are reported in Figure~\ref{fig:setup_regimes}.
We provide 3 possible set-points to a wind turbine $\turbine$, i.e., $\setpoint_\turbine \in \{1490\text{ kW}, 6420\text{ kW}, 8000\text{ kW}\}$ (equivalent to measured wind speeds of 6.5, 10.0 or 13.5 m/s at the turbine's location), which translates into a low, medium and high power production.

\begin{minipage}{\textwidth}
We perform experiments for all combinations of the following parameters:
\begin{itemize}
\item Wind direction: $d \in \{0^\circ, 30^\circ\}$
\item Demand: $\power_\text{dem} \in  \{60\text{ MW}, 70\text{ MW}, 80\text{ MW}, 90\text{ MW}, 100\text{ MW}\}$
\item Number of high-risk turbines: $n \in \{1,2,3,4\}$
\item Penalty:
$\load^\turbine\left(\localsetpoint\right) = \begin{cases}
+\infty\text{ if $\turbine$ is high-risk and }\\
\qquad\power^\turbine_t\left(\localsetpoint\right) \ge 5.2\text{ MW}\\
0 \text{ otherwise}
\end{cases}$
\end{itemize}
\end{minipage}
An infinite penalty is provided to a high-risk turbine when the used set-point leads to damage accumulation, which occurs at 65\% of the maximum power level \cite{alvarez2018improved}. Although any non-linear penalty function can be used, this infinite penalty allows the wind farm operator to identify high-risk turbines and ensure that no high-load set-points are assigned to them (see Section~\ref{sec:challenges}). The $n$ high-risk turbines are randomly chosen. We set the standard deviation $\sigma$ in the prior distribution (Equation~\ref{eq:prior}) to 1 MW, which allows for a sufficient amount of exploration over the complete power range of $[0\text{ MW}, 8\text{ MW}]$.
Each experiment is repeated 100 times.\footnote{The source code of all experiments is publicly available at \url{https://github.com/timo-verstraeten/spts-experiments}.}

We compare SPTS with a set-point allocation strategy, based on the heuristic proposed by \cite{siniscalchi2019wind}. Specifically, higher set-points are first assigned to turbines which are further back in the farm, with respect to the incoming wind vector. This process is repeated toward the front of the farm until the required demand is reached. The solution with the power production that is closest to the demand is recorded.
For both the heuristic and SPTS, the best performing control strategies of each run are compared. As the main focus is to prevent high-load actions on high-risk turbines, we define the performance of a set-point configuration in terms of the total penalty first, and in case of draws, the configuration that matches the demand the closest is chosen.

\begin{figure}[!h]
 \begin{subfigure}{0.49\textwidth}
 \centering
\includegraphics[scale=0.4,trim=0.7cm 0cm 0cm 0cm]{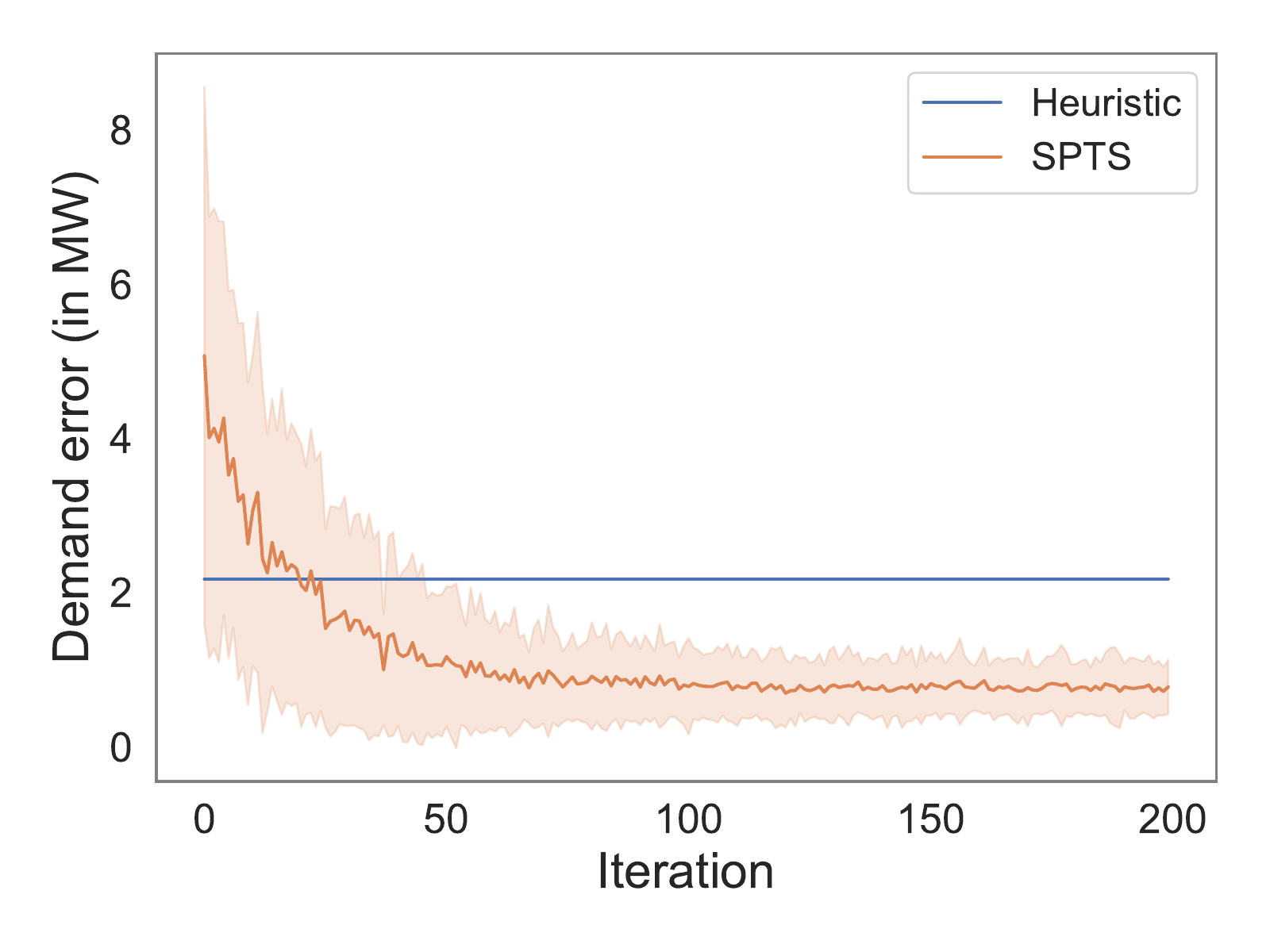}
\caption{Demand error}
\end{subfigure}
 \begin{subfigure}{0.49\textwidth}
 \centering
\includegraphics[scale=0.4,trim=0cm 0cm 0cm 0cm]{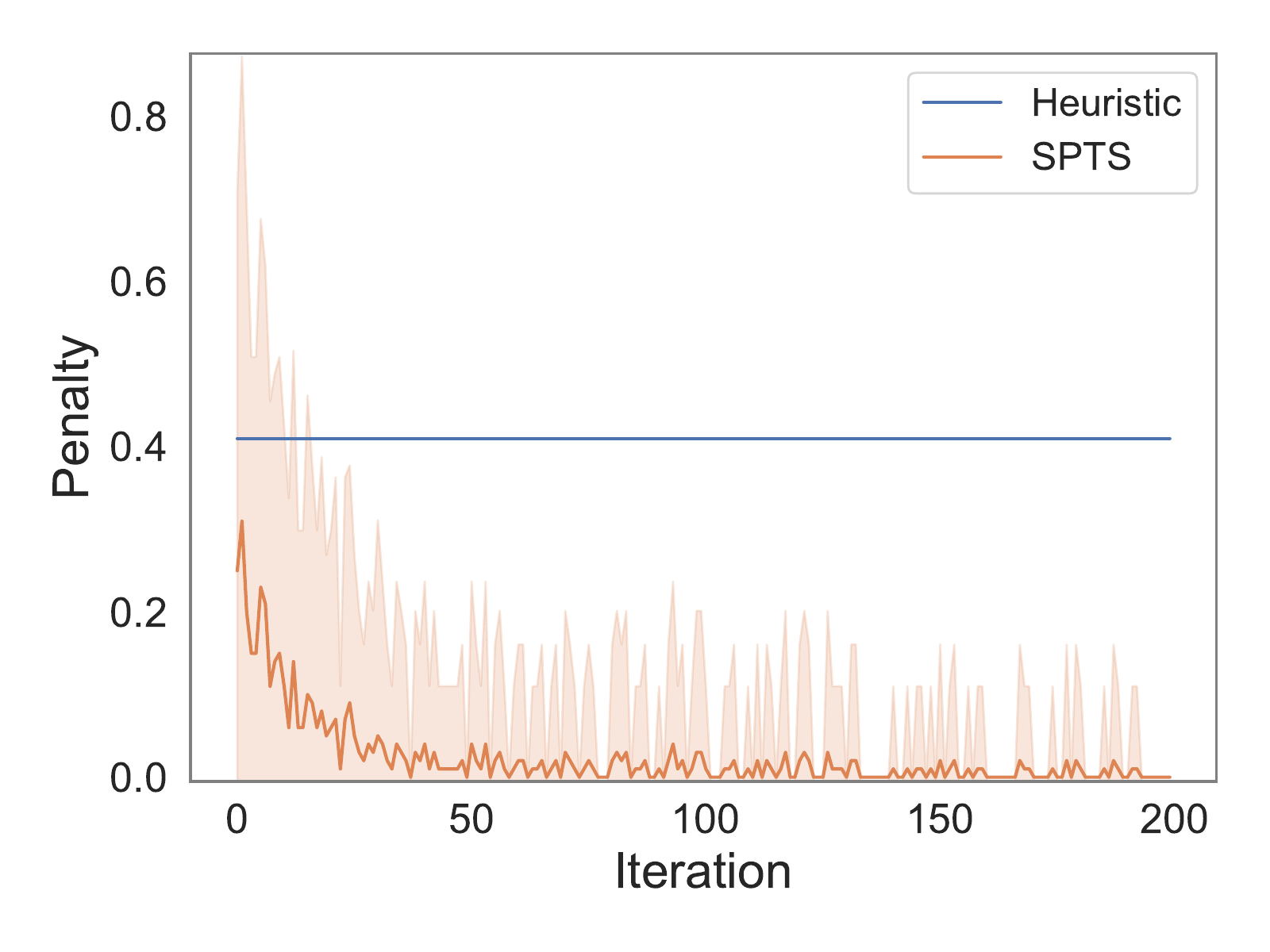}
\caption{Penalty}
\end{subfigure}
\caption{Learning curves of the demand error (a) and total penalty (b) obtained at each iteration, for a wind direction of $0^\circ$, a demand of $80$ MW and $3$ high-risk turbines. The average trend (line) and standard deviation (shaded area) are shown. The experiment is repeated 100 times.}
\label{fig:learning_curve}
\end{figure}

Figure~\ref{fig:learning_curve} shows the learning curve of SPTS for the setting with a wind direction of $0^\circ$, a demand of $80$ MW and $3$ high-risk turbines. The trend indicates that 200 iterations are sufficient to ensure convergence. The learning curves for all settings are reported in Appendix~A. Note that the heuristic is a deterministic approach, and thus the variance on the outcomes, over multiple repetitions of the experiment, is zero.

\begin{figure*}[!t]
\centering
 \begin{subfigure}{0.49\textwidth}
\centering
\includegraphics[scale=0.35,trim=1cm 0cm 0cm 0cm]{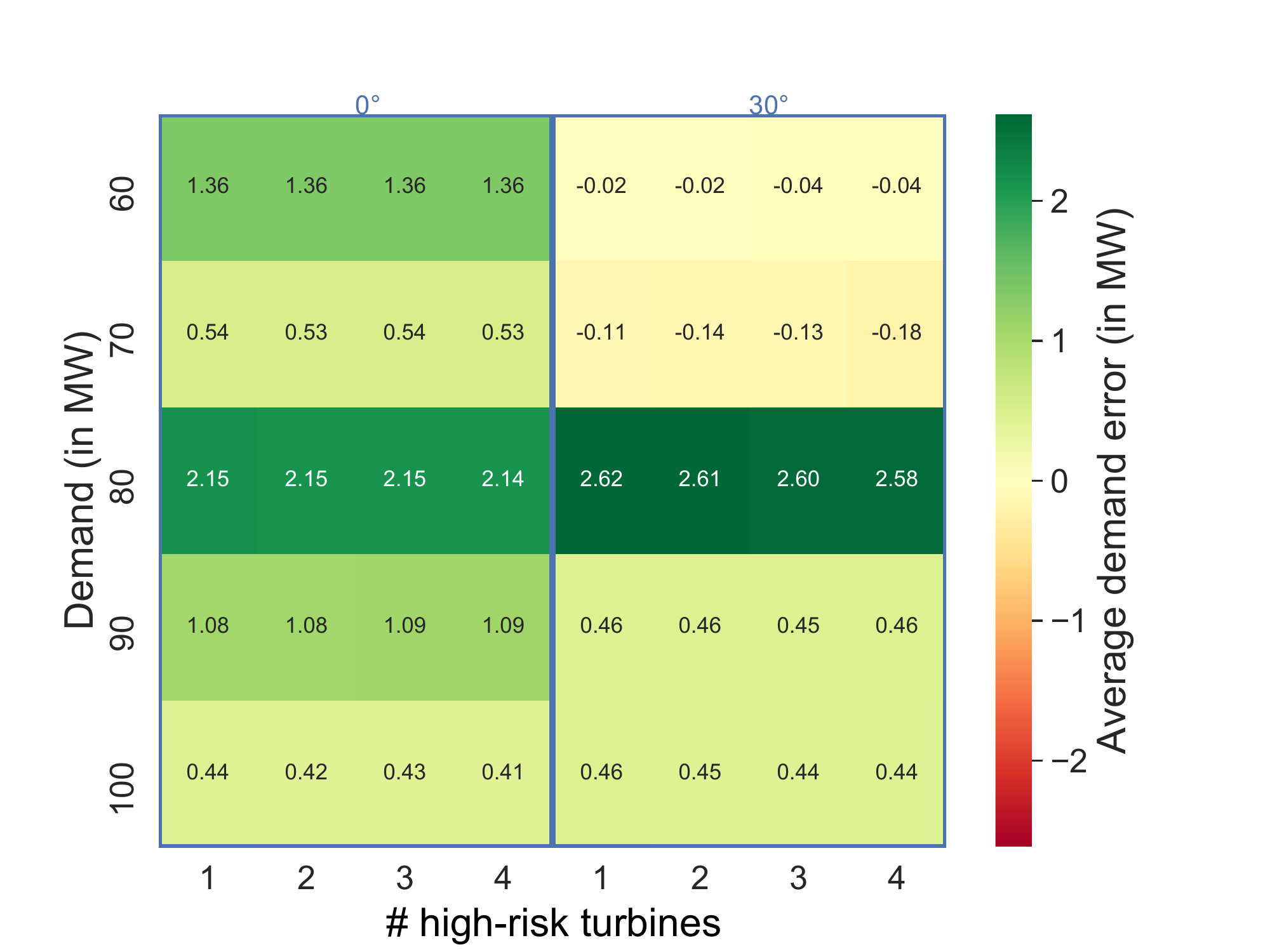}
\caption{Demand error}
\label{fig:heatmap_error}
\end{subfigure}
 \begin{subfigure}{0.49\textwidth}
\centering
\includegraphics[scale=0.35,trim=0cm 0cm 0cm 0cm]{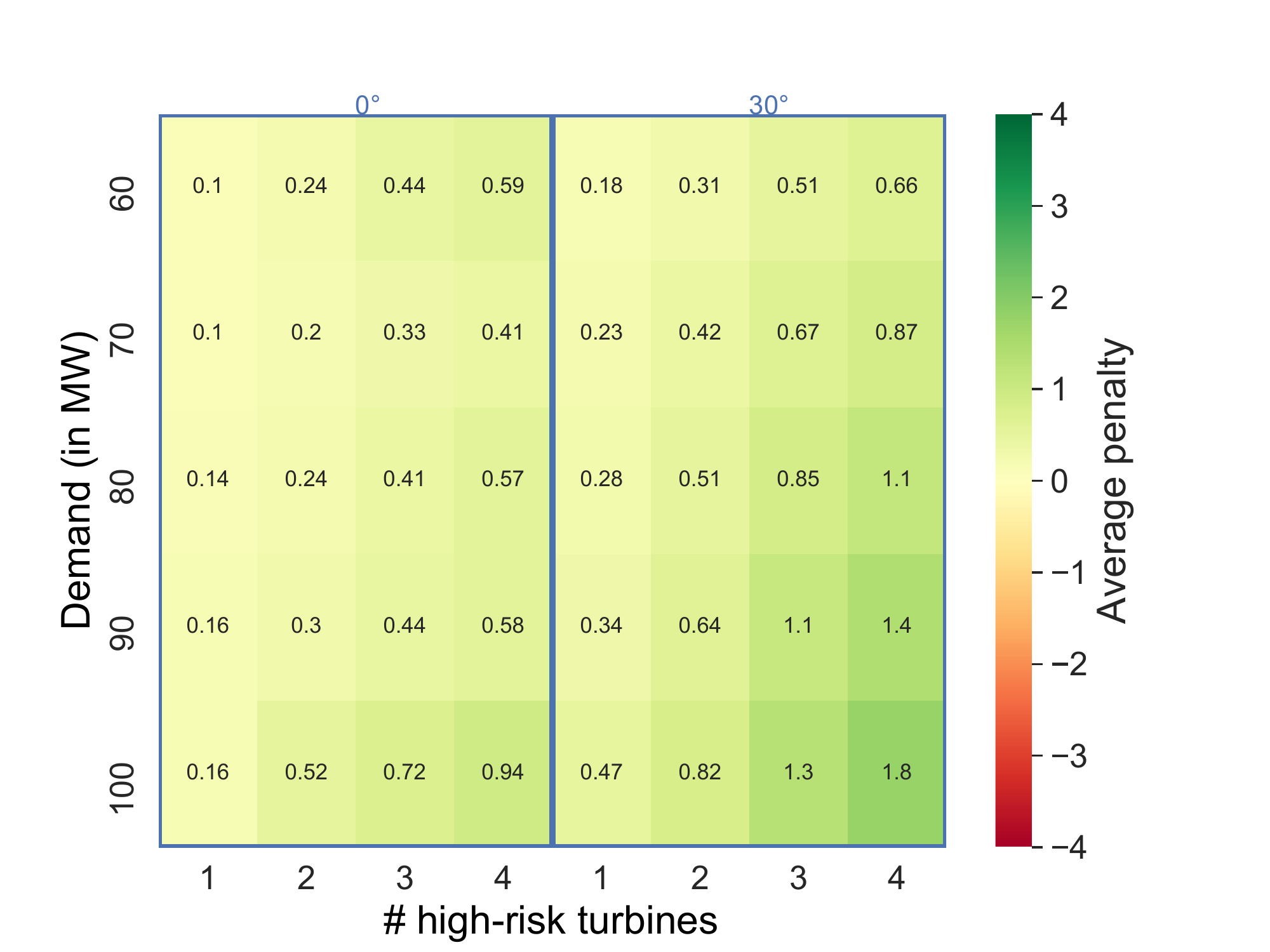}
\caption{Penalty}
\label{fig:heatmap_penalty}
\end{subfigure}
\caption{Heatmaps of the average absolute difference between the performance of the heuristic and of SPTS. The performance in terms of demand error (left) and total penalty (right) are plotted.
A positive value (green) indicates a better average performance obtained by SPTS, while a negative value (red) indicates a worse performance compared to the heuristic. The performances are averaged over 100 samples.}
\label{fig:heatmaps}
\end{figure*}

Figure~\ref{fig:heatmaps} shows the average absolute difference between the best performances of the heuristic and of SPTS for all parameter combinations, both in terms of penalty and demand error. On the one hand, SPTS reaches comparable or better results than the heuristic in terms of demand error. On the other hand, the heuristic approach receives more penalties when the number of high-risk turbines is increased, or when the required demand is increased. This is expected, as the heuristic approach will allocate higher set-points to high-risk machines to reach a higher demand. Moreover, SPTS significantly outperforms the heuristic in terms of demand error when the demand is 80 MW. This is due to the fact that many possible set-point configurations exist to meet this demand, and can thus easily be found by SPTS. In contrast, when the demand is 60 MW or 100 MW, only a few configurations are viable, in which most of the set-points are low or high, respectively.
It is important to note that, over all runs, the best configurations achieved by SPTS \emph{never} contained a damage-inducing set-point assigned to a high-risk turbine (i.e., the total penalty of the solution is always 0 for all settings). 
Box plots of the best performing control strategies obtained by SPTS and the heuristic for all parameter combinations are reported in Appendix~B.

\begin{figure}[!h]
 \begin{subfigure}{0.49\textwidth}
 \centering
\includegraphics[scale=0.3,trim=0cm 0cm 0cm 0cm,clip]{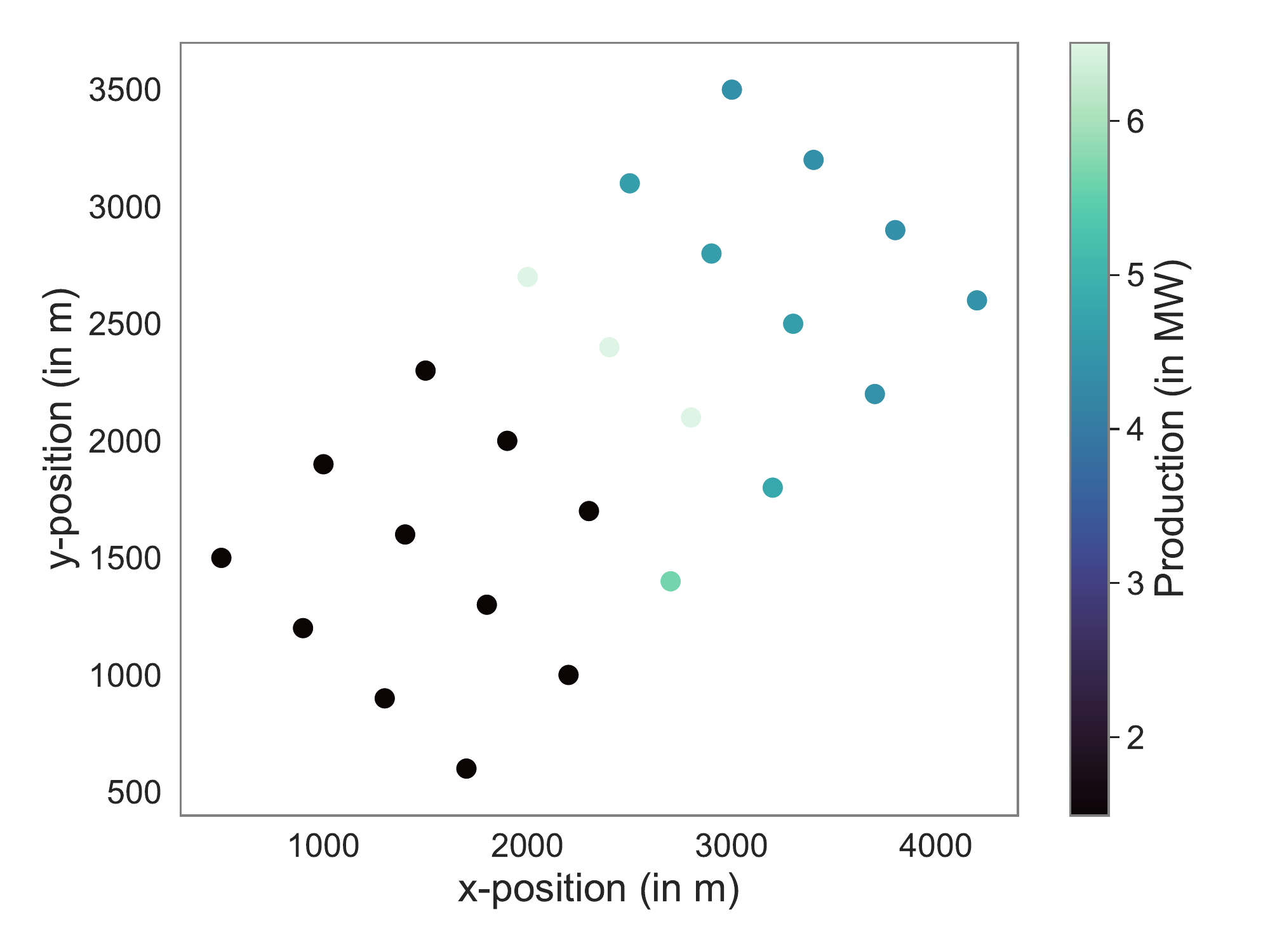}
\caption{Heuristic}
\end{subfigure}
 \begin{subfigure}{0.49\textwidth}
 \centering
\includegraphics[scale=0.3,trim=0cm 0cm 0cm 0cm,clip]{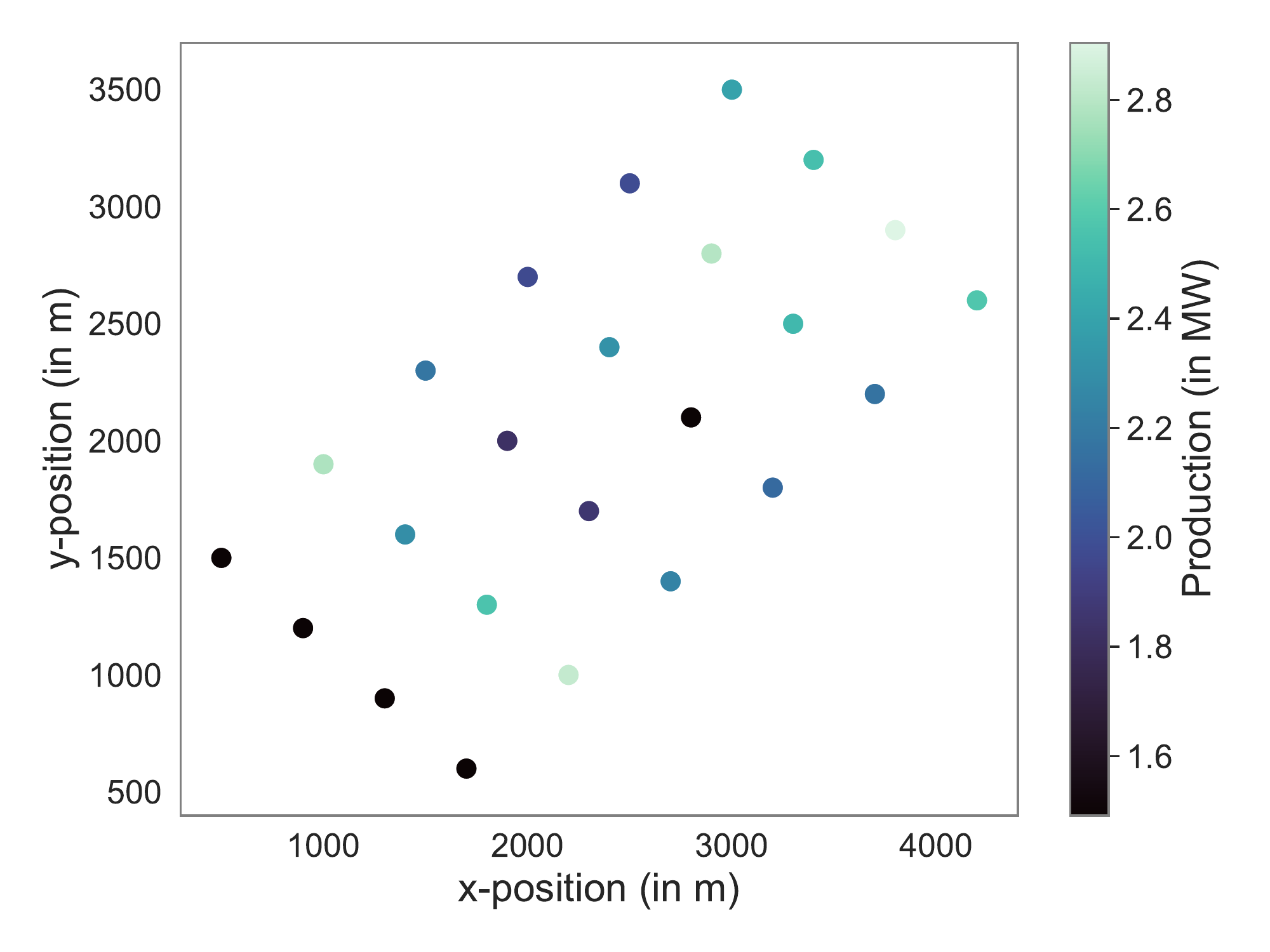}
\caption{SPTS}
\end{subfigure}
\caption{Farm-wide view of the power productions for the wind direction of $0^\circ$, a demand of $80$ MW and $3$ high-risk turbines. The power productions for the best set-point configurations found by the heuristic approach (a) and \methodacronym\ (b) are shown. The power productions are averaged over $100$ runs.}
\label{fig:farm_power}
\end{figure}

Figure~\ref{fig:farm_power} provides a farm-wide view of the power productions obtained, given the best set-point configuration found by the heuristic approach and SPTS. This is shown for the setting with a wind direction of  $0^\circ$, a demand of 80 MW and 3 high-risk turbines. The farm-wide power productions for all settings are reported in Appendix~C.
By the heuristic's definition, set-points are first maximized for downstream turbines. This process is repeated toward the front of the farm. Therefore, it is expected to see a pattern in which the downstream turbines exhibit a higher actual power production, compared to the upstream turbines. While the heuristic generates an intuitive pattern, the solution of \methodacronym\ indicates that this is not necessarily an optimal solution. Therefore, the results shown in Figure~\ref{fig:heatmap_error} and~\ref{fig:farm_power} demonstrate that a data-driven approach is essential in order to come up with context-aware solutions.
Nevertheless, due to the allocation of demand over the different regimes, \methodacronym\ also favors higher set-points for downstream turbines. The reduced power generation of turbines with higher observed loads, combined with the ability to include arbitrary penalty functions (see Figure~\ref{fig:heatmap_penalty}), demonstrates that \methodacronym\ is suitable to match the power demand in a manner that reduces overall lifetime consumption.

\section{Discussion}
\label{sec:discussion}

We propose a new wind farm control algorithm that allocates set-points taking into account load information. The results demonstrate that \methodacronym\ can successfully match the power demand without assigning high-load set-points to turbines.

We focus on a wind farm with a symmetric grid-like structure, which comprises the majority of contemporary wind farm designs. However, our method is not biased toward symmetric topologies. Instead, \methodacronym\ guides the learning process by using information about the environmental and operational conditions at the farm site, acquired from wake analyses and knowledge regarding the turbines' health. Therefore, \methodacronym\ is applicable to irregularly shaped wind farms, which can be constructed due to landscape constraints \cite{gu2013irregular}.

To model the turbine's dynamics and wind flow, we use the FLORIS wake simulator, in which set-point configurations can be evaluated quickly. \methodacronym\ has the ability to include expert knowledge without being dependent on the simulator used. Therefore, it is easy to switch between different types of simulators, such as computational fluid dynamics simulators \cite{castellani2013practical}.

In our experiments, we focused on optimizing control strategies under steady wind conditions. Therefore, it is sufficient to analyze the expected performance of control strategies in a noise-free setting. Nevertheless, in the case of transient wind conditions (e.g., wind gusts or storms), it may be important to investigate the variance on the performance measure as well. Since \methodacronym\ uses a Bayesian sampling approach, it is straightforward to introduce a likelihood distribution with a (possibly unknown) noise parameter and update the posterior distributions according to Bayes' rule \cite{russo2017tutorial}.

Our approach uses multi-agent Thompson sampling (MATS) for sampling the set-point configuration space. For MATS, an asymptotic upper bound was established on the cumulative regret, i.e., the total performance loss obtained by executing sub-optimal actions during the learning phase. As we investigated a noise-free setting, every set-point configuration will likely be executed once eventually, and thus the optimal set-point configuration will be found. Therefore, the asymptotic upper bound has no practical use here. Nevertheless, the composition of the bound suggests that the performance loss of \methodacronym\ is in terms of the number of local joint set-points, rather than the number of global joint set-points, as is the case in traditional TS. Therefore, it is expected that \methodacronym\ can learn efficiently using noisy set-point evaluations as well.

Finally, SPTS is a sampling technique inspired by TS to select promising joint actions. Recently, theoretical guarantees have been established for TS in terms of cumulative regret, i.e., the total difference between the expected reward of the optimal (unknown) action and chosen actions obtained \emph{during} the learning process \cite{agrawal2012analysis}.  However, as the set-point optimization is performed \emph{in silico}, the performance of the evaluated set-points during the learning phase is only used to guide the learning process. In our setting, we focus on the performance of the end result, i.e., the best set-point configuration obtained \emph{after} learning. 
This highlights the distinction between exploration-exploitation and best arm identification \cite{audibert2010best}. Note that this is a difference in terms of convergence speed, rather than optimality, as the action chosen by TS still converges to the optimal one (under certain conditions) \cite{lattimore_szepesvari_2020}.
To our knowledge, there are no Bayesian best arm identification algorithms available for dealing with factored multi-agent systems. Although the learning curves converge quickly (see Supplementary Material), further research into best arm identification algorithms for loosely-coupled multi-agents is warranted to improve sample efficiency.

\section{Future Challenges}
\label{sec:challenges}

In our experiments, we use objective functions that heavily penalize a set of turbines, which allows wind farm operators to mark high-risk turbines and reduce their loading conditions. Still, \methodacronym\ finds the optimal joint set-point configuration under arbitrary penalties (see Section~\ref{sec:problem}). Such flexibility is required for capturing the multi-dimensional load spectrum that is present in wind turbine technology. However, further research needs to be conducted to establish the links between turbine responses during dynamic events and potential failure modes \cite{load_causes_2016,load_failure_2017,verstraeten2019fleetwide}.
Through fundamental research and condition monitoring analyses, maintenance costs can be formalized as a penalty function within \methodacronym, which is important toward the further development of advanced wind farm controllers.

\methodacronym\ finds the optimal combination of set-points taken from a discrete space. To further reduce the demand error, continuous set-points should be considered. Optimizing over a factored representation in continuous space is challenging, as the choice of a turbine is possibly dependent on an infinite amount of configurations chosen by its parents. To our knowledge, no optimization algorithms currently exist that fully operate within a factored continuous action space. Therefore, research should focus on continuous optimization techniques for loosely-coupled multi-agent systems, such that accurate solutions can be provided in a feasible manner. Nevertheless, one can use the optimal discrete set-point configuration provided by \methodacronym\ as a starting point and further optimize over the continuous set-point space using an iterative approach \cite{siniscalchi2019wind}. Naturally, such a solution may not be provably optimal and convergence may be challenging to guarantee.

Our approach considers each scenario independently and does not generalize over environmental conditions. In data-driven wind farm control research, control strategies are often learned without considering the dependencies between environmental parameters (e.g., wind speed and wind direction) \cite{vandijk2016,verstraeten2019fleetwide}. However, generalizing over environmental conditions, for instance, through the use of contextual bandits \cite{agrawal2013thompson}, would improve the sample-efficiency of \methodacronym\ over multiple settings.

Finally, to accurately represent the current environmental conditions at the real wind farm, a mechanism is necessary to reduce measurement noise, while retaining the fine granularity of the data sources. To this end, a similarity-based data exchange between the turbines can be used to process real wind farm data \cite{fleet_gprl_2020} and provide a reliable parametrization to the wake simulator. Combined with the \methodacronym\ control algorithm, scenario optimization can be performed to provide set-point configurations in real time, allowing \methodacronym\ to be used in the real world.

\section{Conclusion}
We proposed Set-Point Thompson sampling (\methodacronym), a novel Bayesian algorithm for wind farm control. The method exploits the sparse topological structure of wind farms to efficiently search for the optimal set-point configuration. We showed that \methodacronym\ achieves a comparable or better performance than the heuristic approach in terms of demand error. Yet, \methodacronym\ manages to consistently assign lower set-points to high-risk turbines, which effectively reduces their lifetime consumption.

The results demonstrate that a data-driven approach is necessary when dealing with complex penalty functions. Complex penalty functions are inevitable when considering the multi-dimensional load spectrum of wind turbine technology. Introducing knowledge about the load spectrum is key to advance the current state-of-the-art in wind farm control, as loads have a large impact on the remaining useful life of the turbines, and ultimately on maintenance costs.

Still, the ability to introduce wind farm properties and mechanical knowledge into the data-driven wind farm controller is necessary to render the learning process tractable. Therefore, closing the gap between physics-based heuristics and data-driven learning will ensure the development of advanced wind farm controllers that are well-balanced in terms of learning tractability, flexibility and optimality.

\section*{Acknowledgments}
The authors would like to acknowledge FWO (Fonds Wetenschappelijk Onderzoek) for their support through the SB grants of Timothy Verstraeten (\#1S47617N) and Eugenio Bargiacchi (\#1SA2820N), and the post-doctoral grant of Pieter J.K. Libin (\#1242021N). This research was supported by funding from the Flemish Government under the ``Onderzoeksprogramma Artifici\"ele Intelligentie (AI) Vlaanderen'' programme and under the VLAIO Supersized 4.0 ICON project.


\clearpage
\bibliographystyle{plain} 
\bibliography{refs}

\section*{Appendix A -- Learning Curves}

\subsection*{A.1 -- Demand Error}
\sptslearncurve{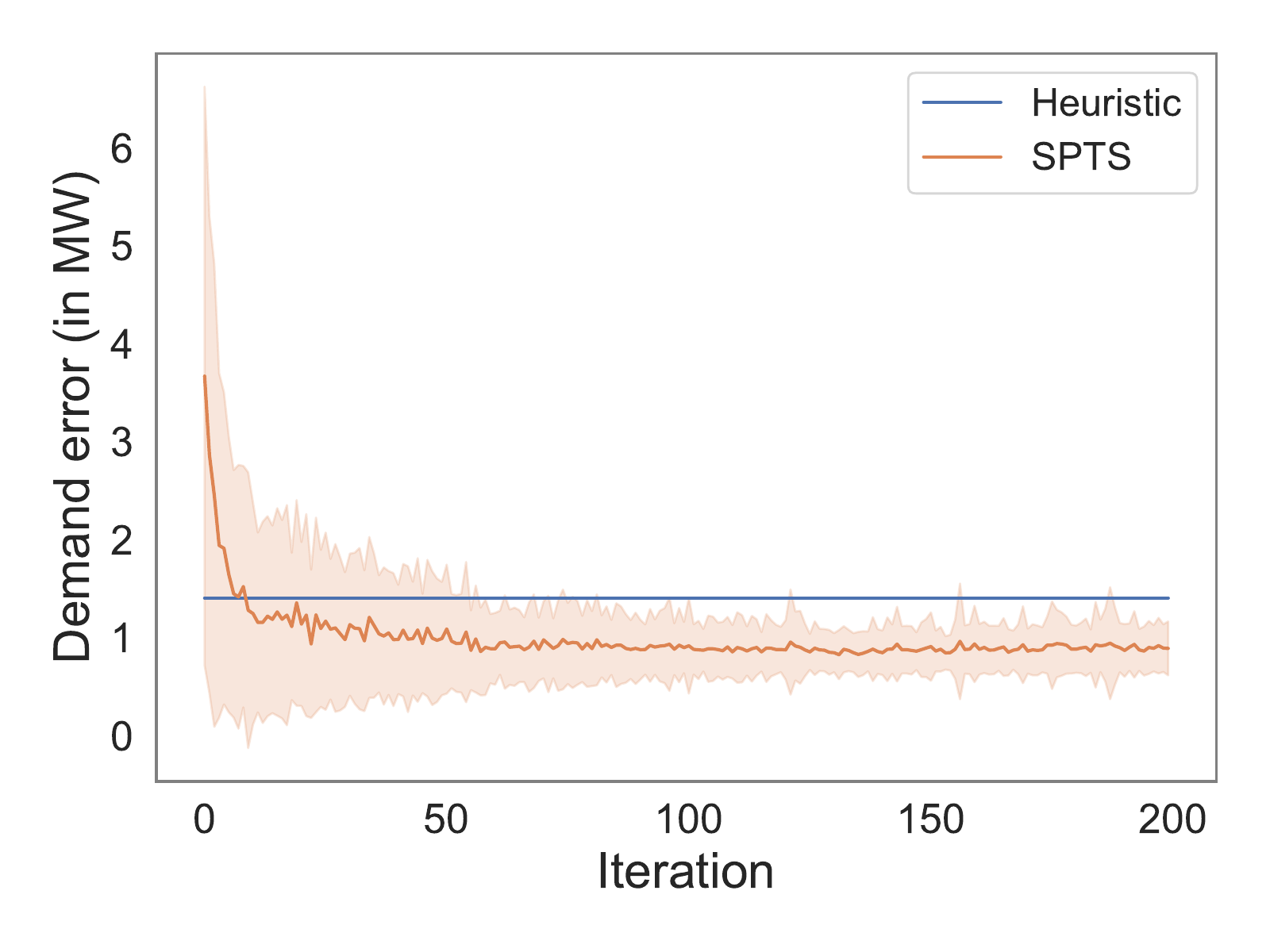}{Demand error}{0}{60}
\sptslearncurve{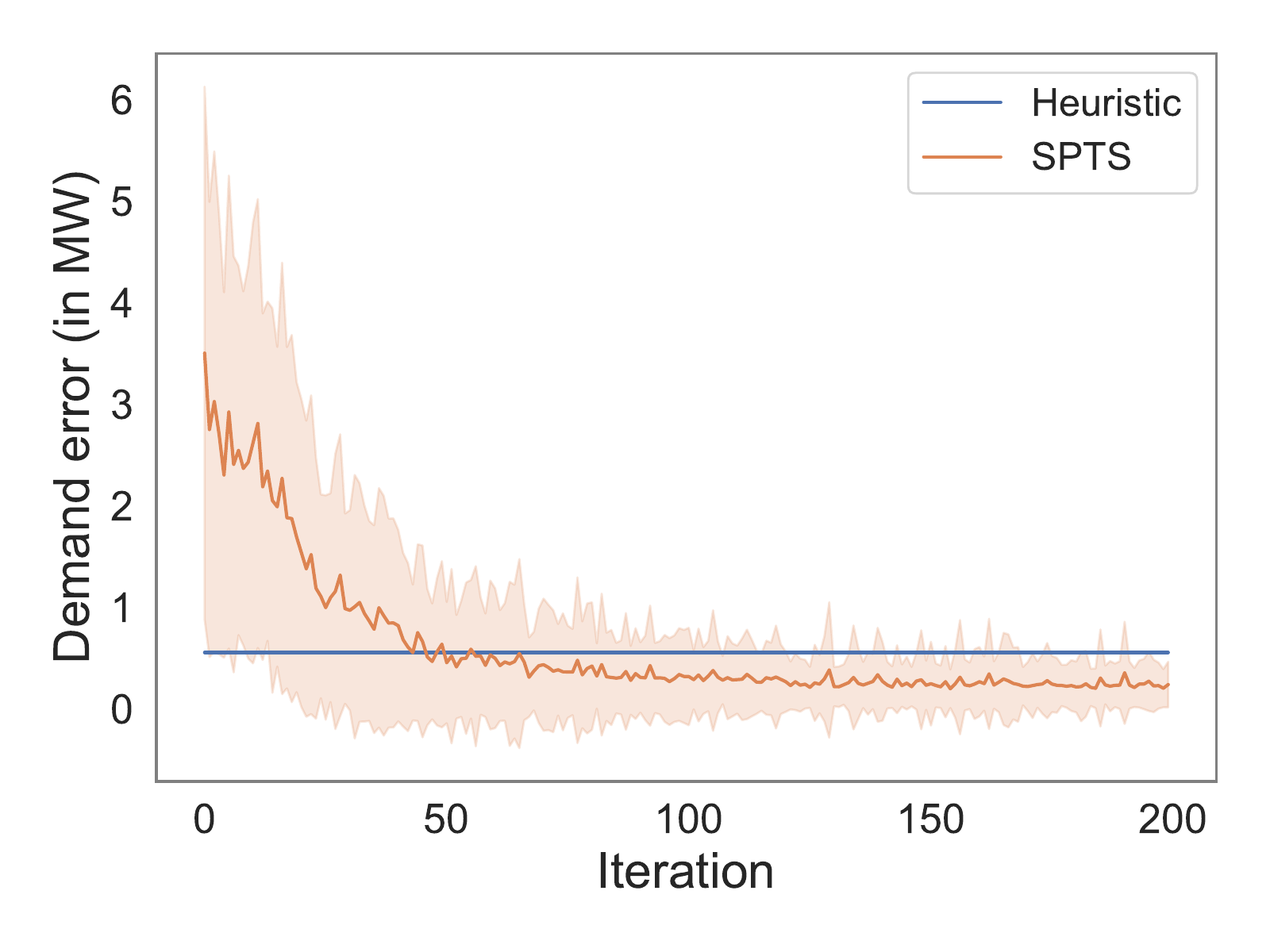}{Demand error}{0}{70}
\sptslearncurve{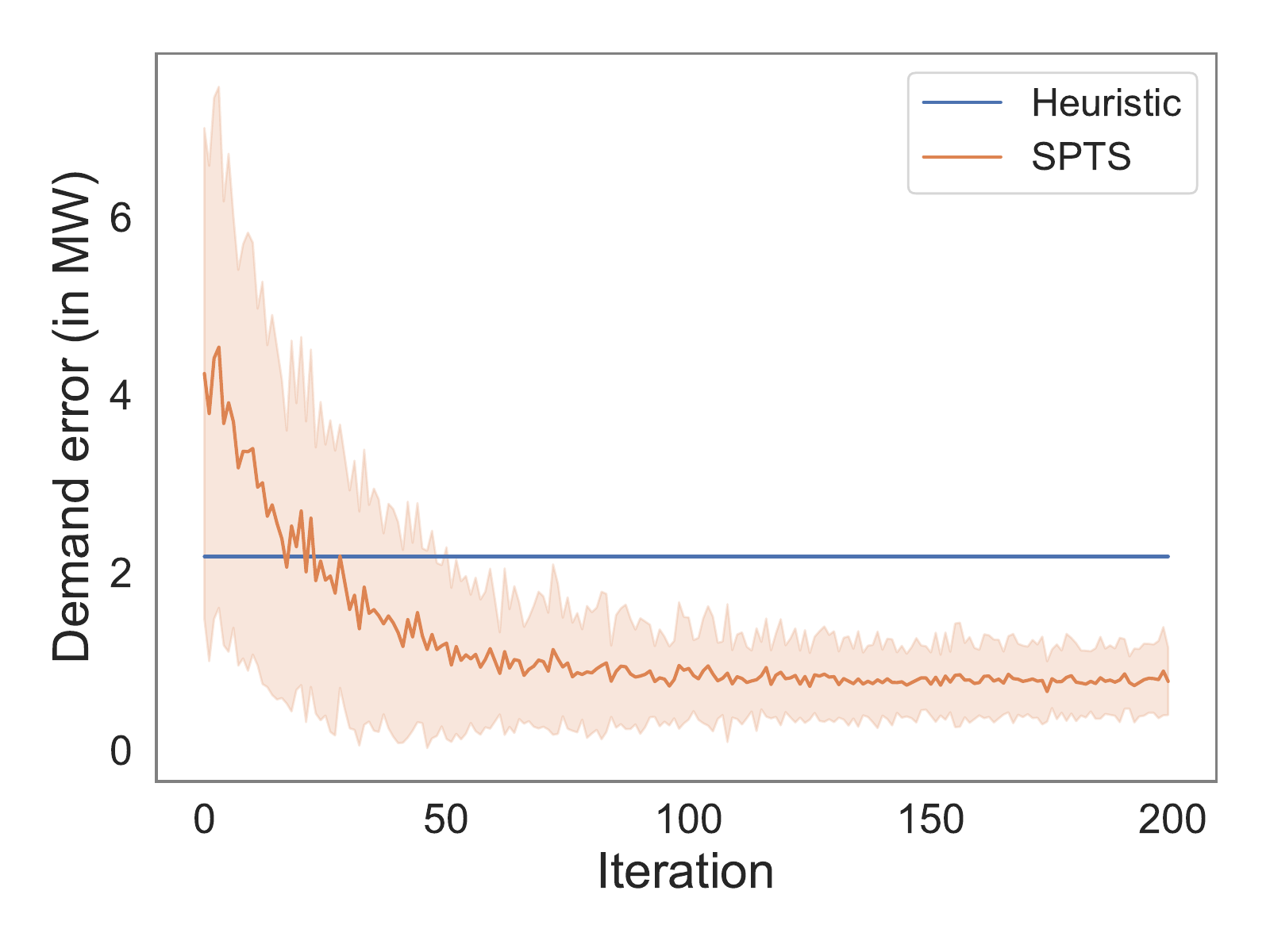}{Demand error}{0}{80}
\sptslearncurve{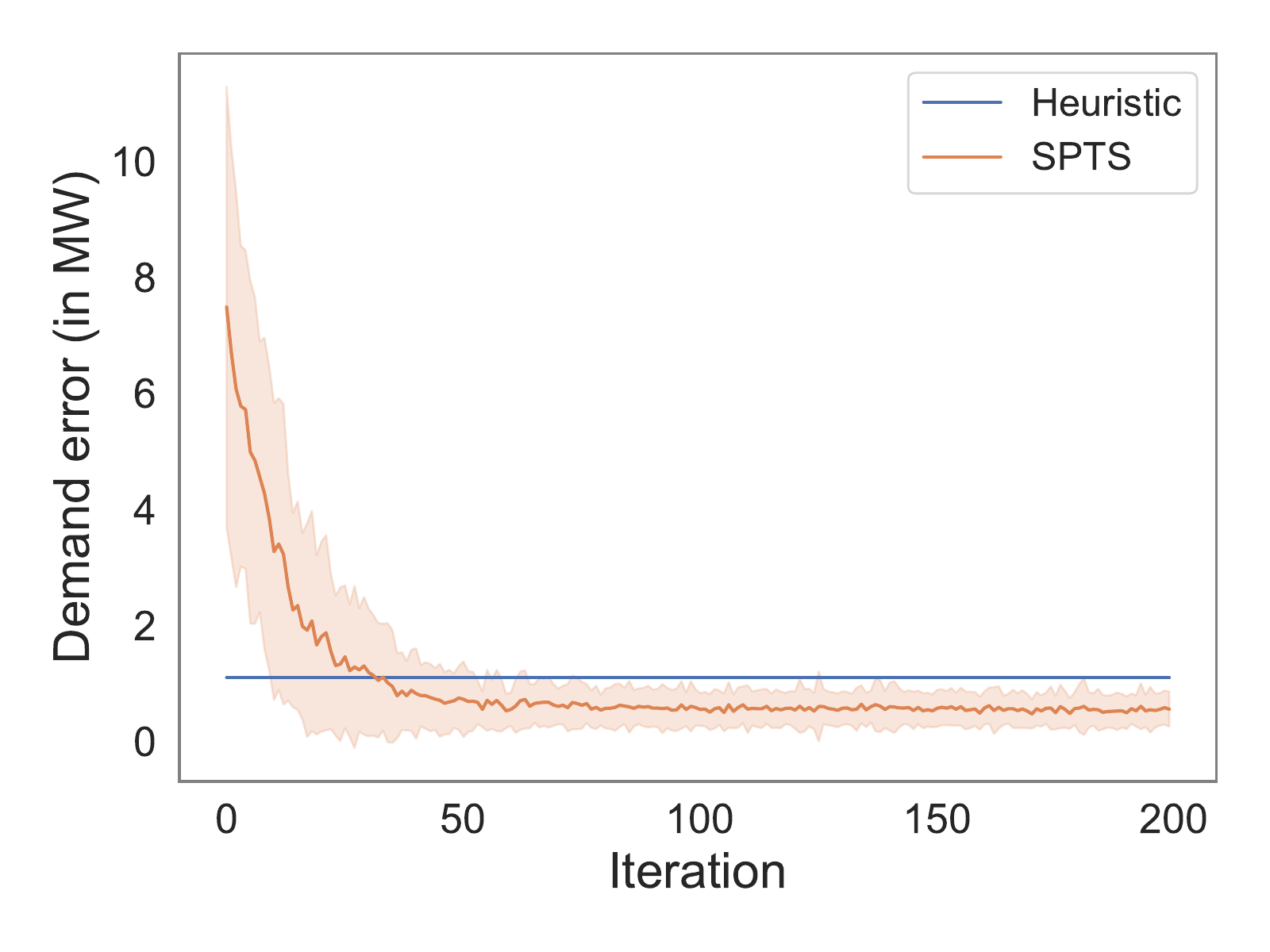}{Demand error}{0}{90}
\sptslearncurve{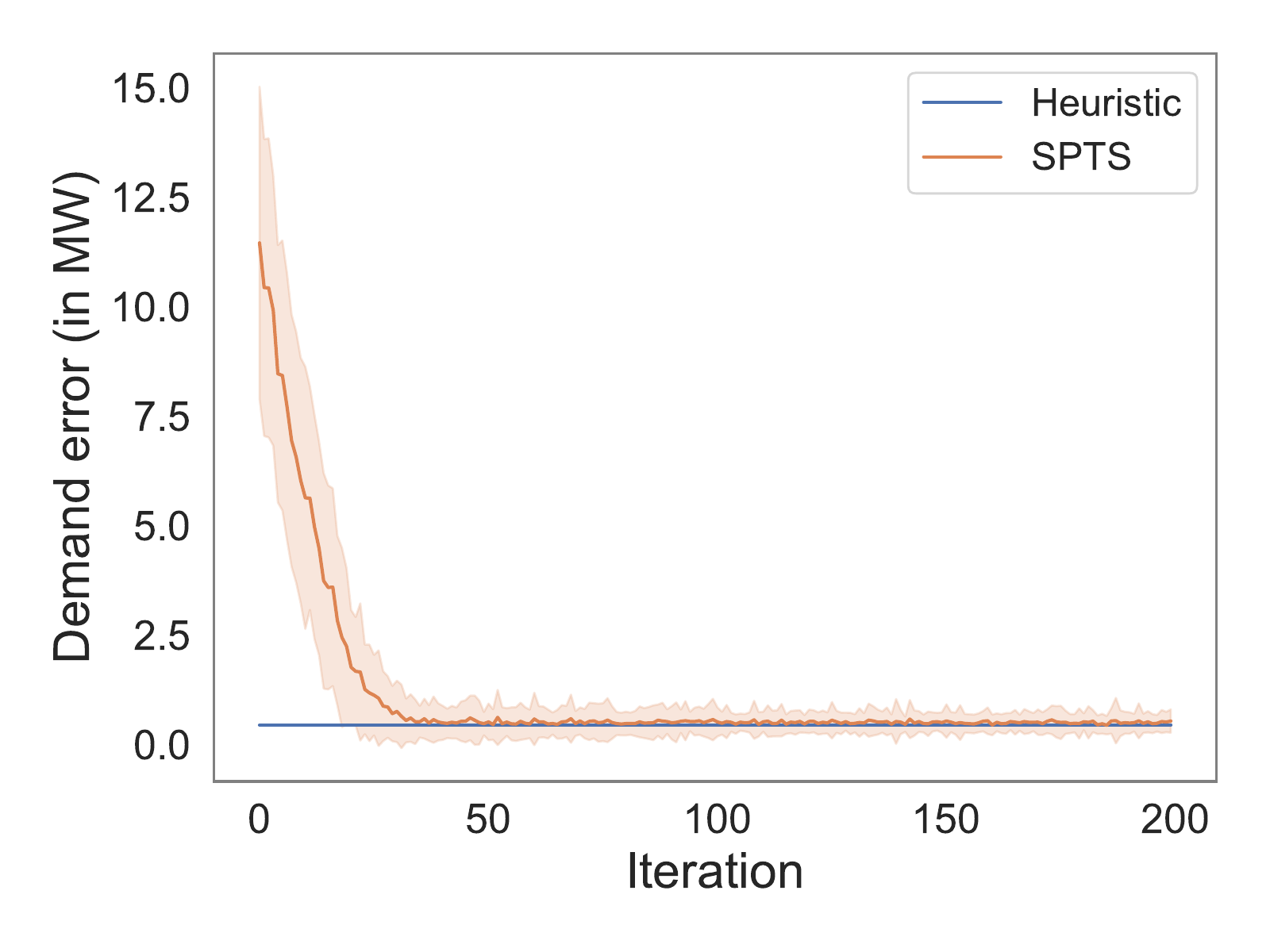}{Demand error}{0}{100}
\sptslearncurve{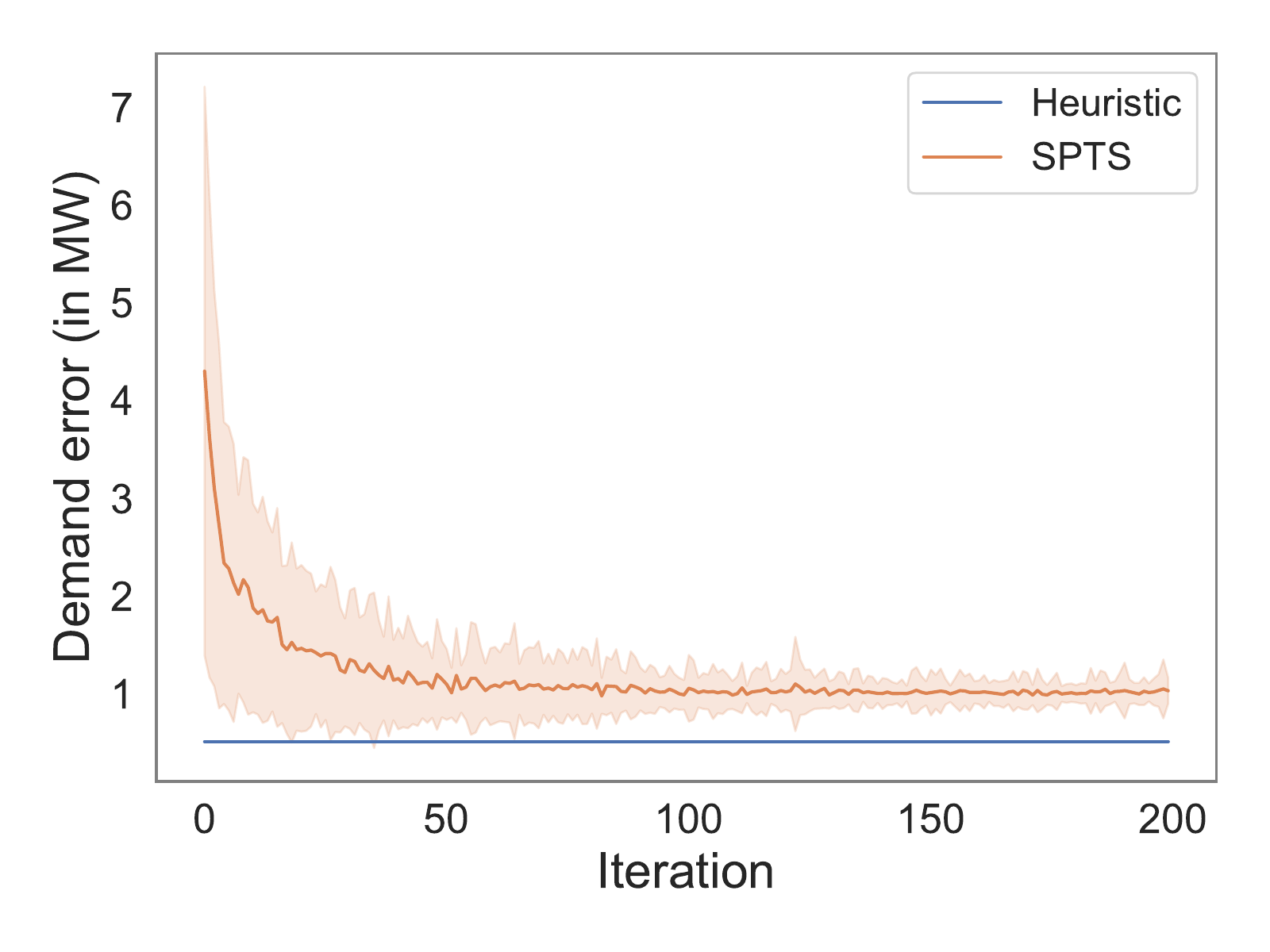}{Demand error}{30}{60}
\sptslearncurve{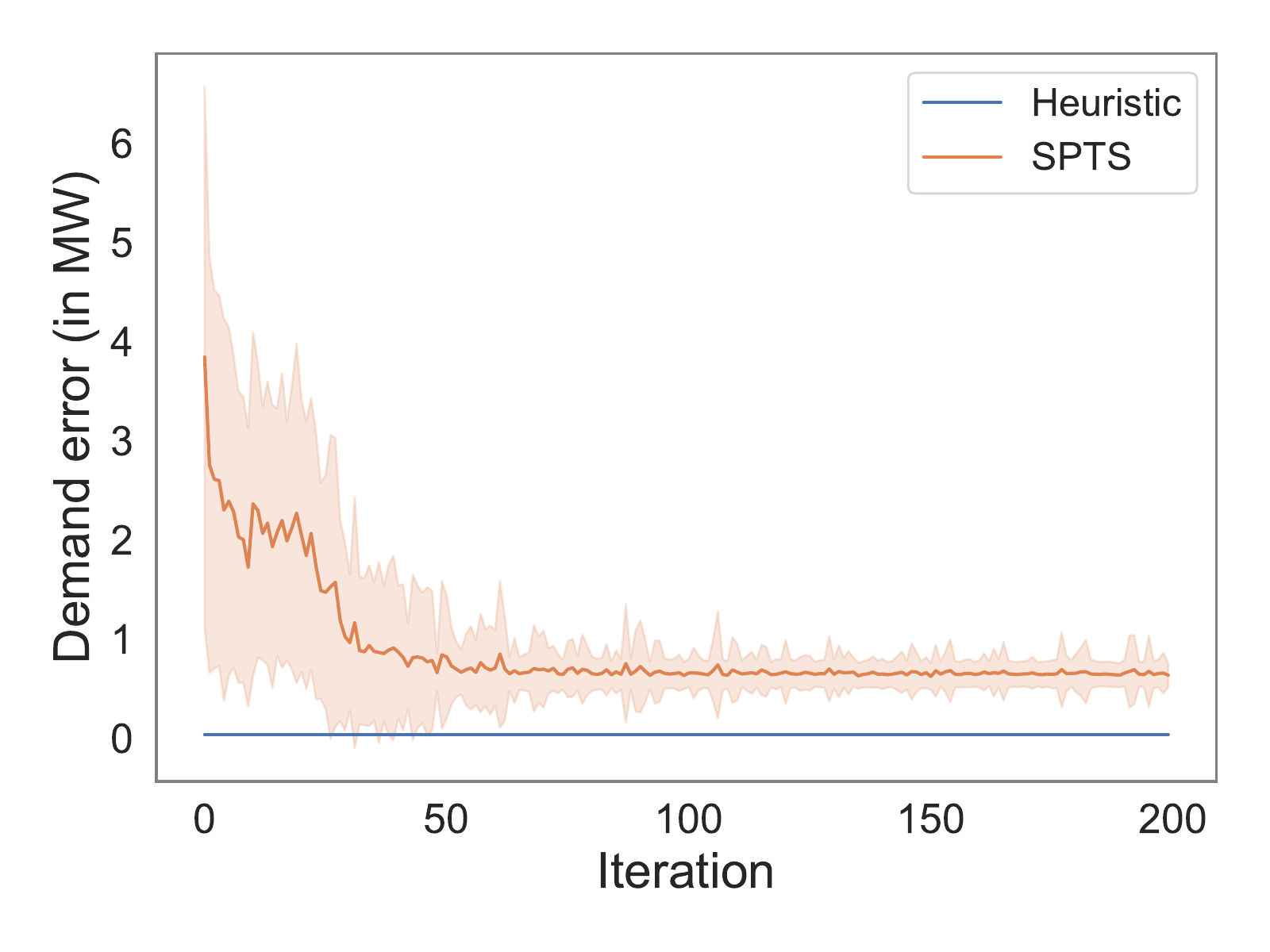}{Demand error}{30}{70}
\sptslearncurve{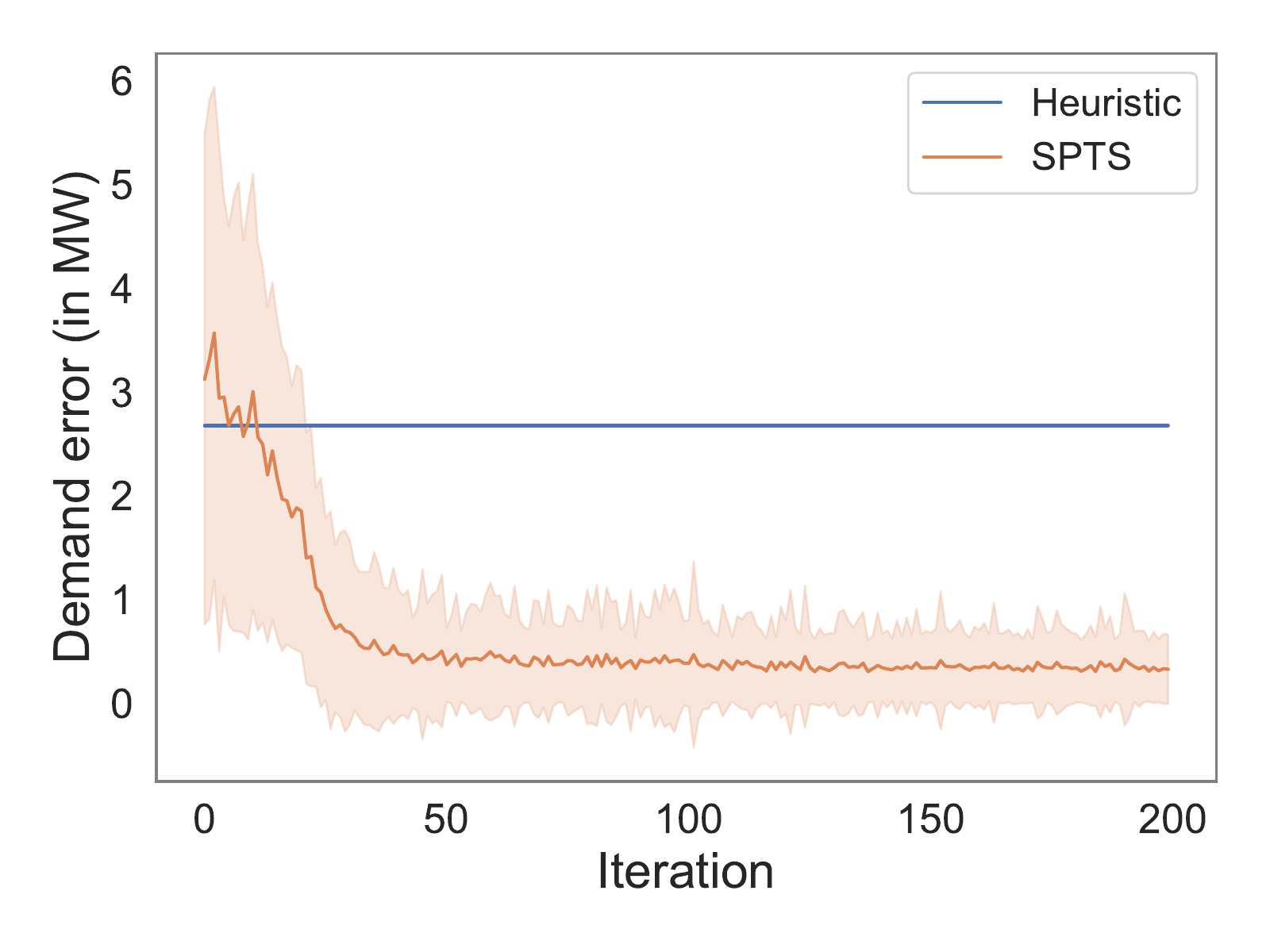}{Demand error}{30}{80}
\sptslearncurve{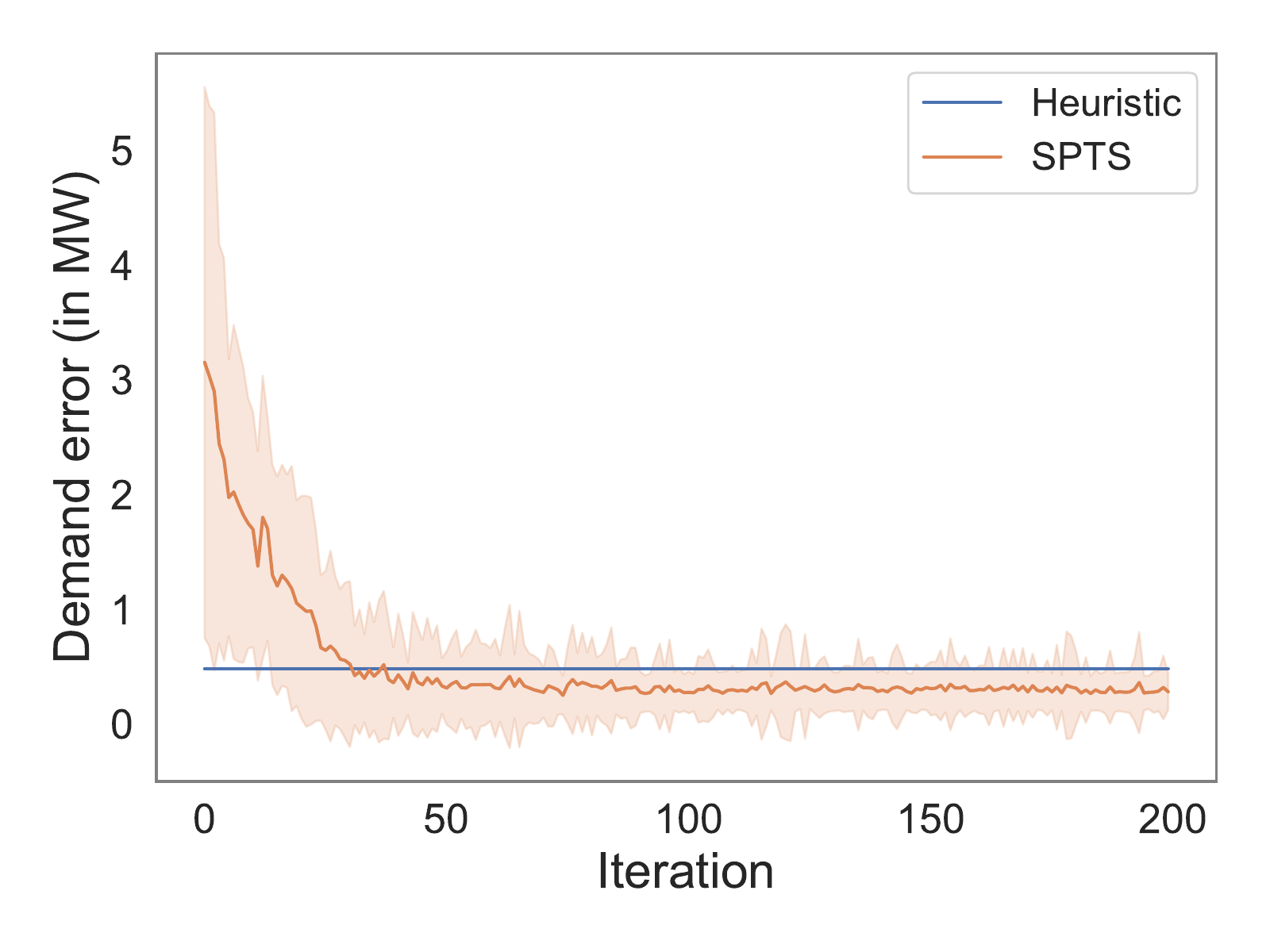}{Demand error}{30}{90}
\sptslearncurve{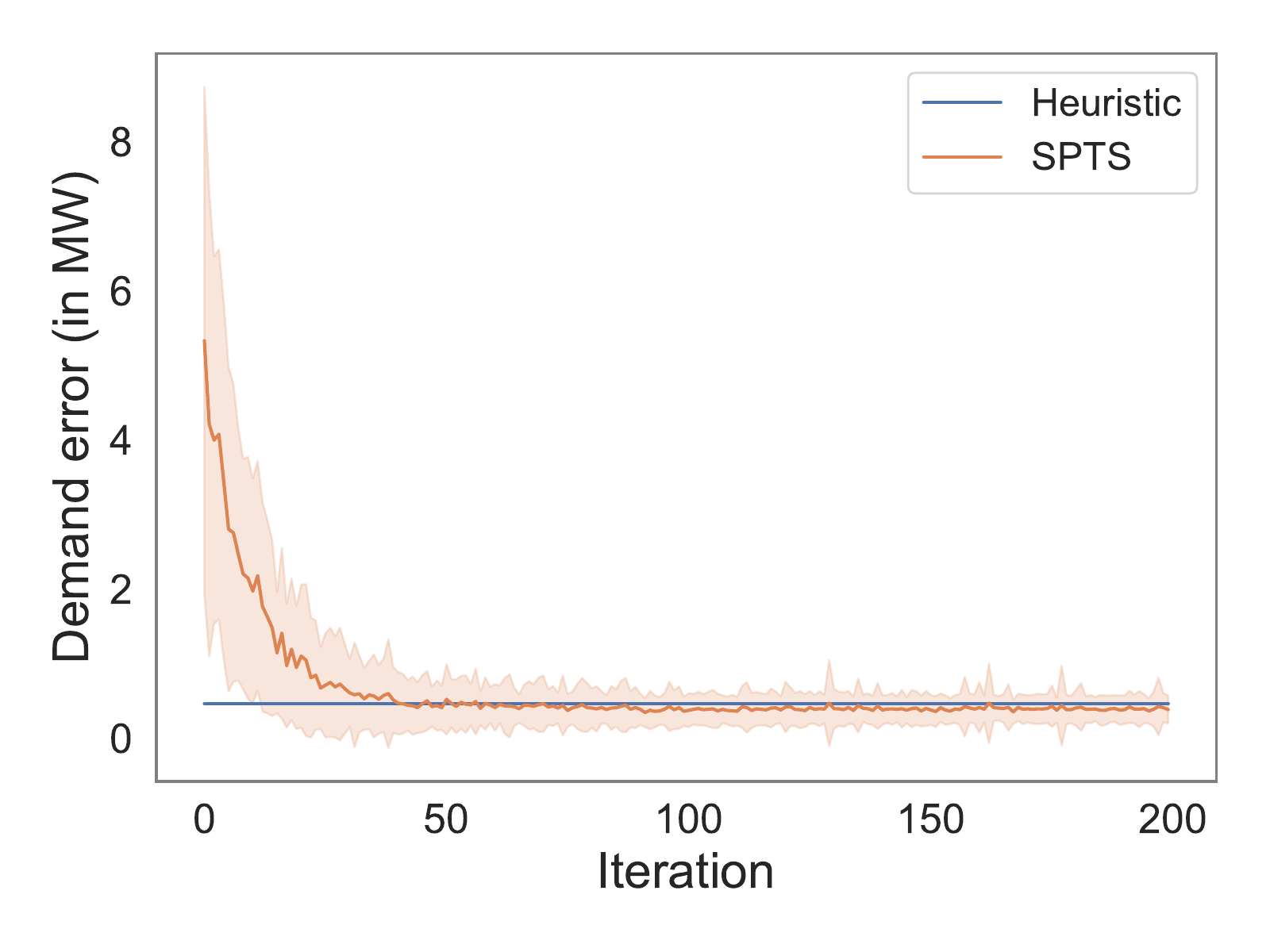}{Demand error}{30}{100}
\clearpage

\subsection*{A.2 -- Penalty}
\sptslearncurve{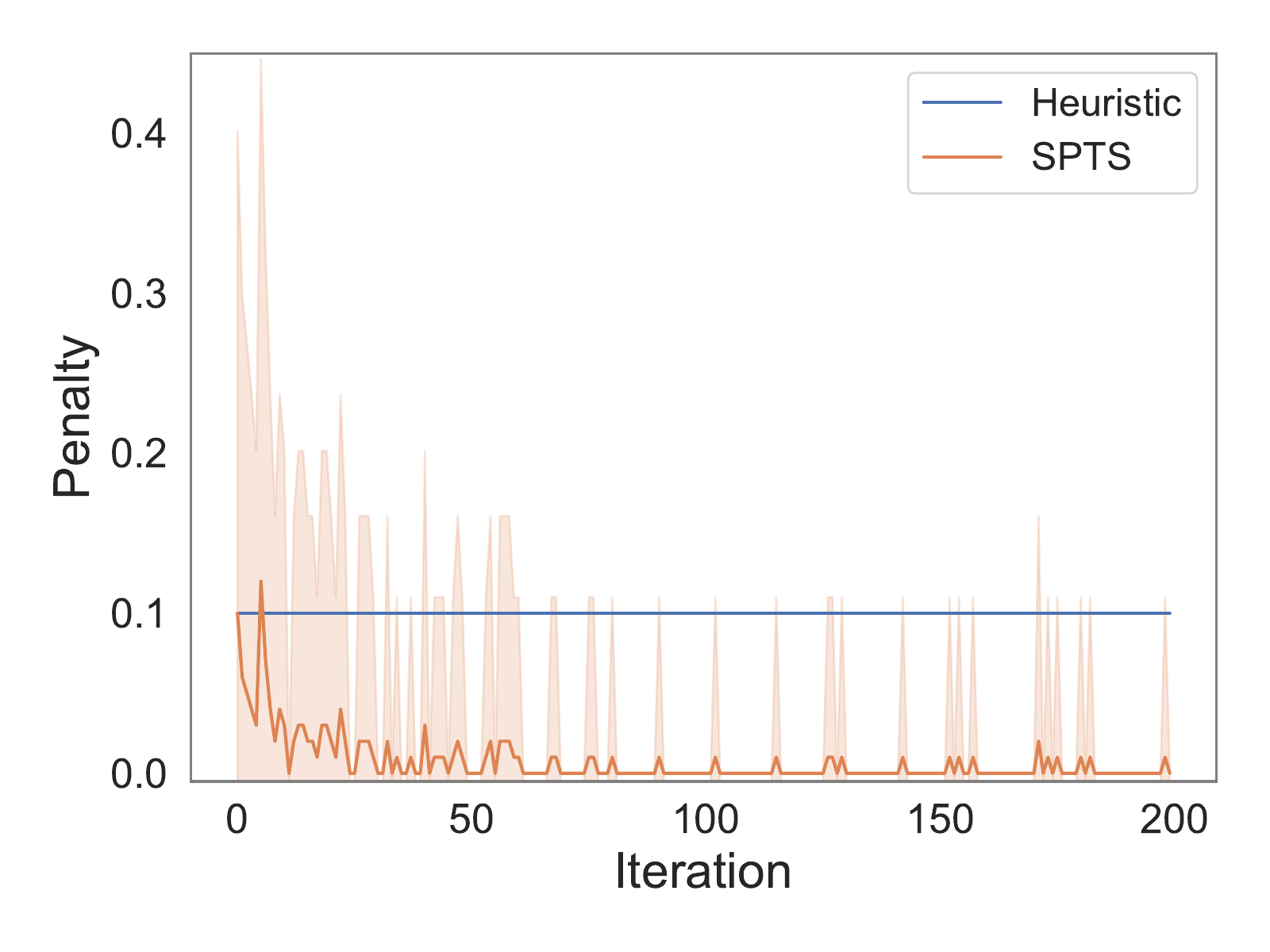}{Penalty}{0}{60}
\sptslearncurve{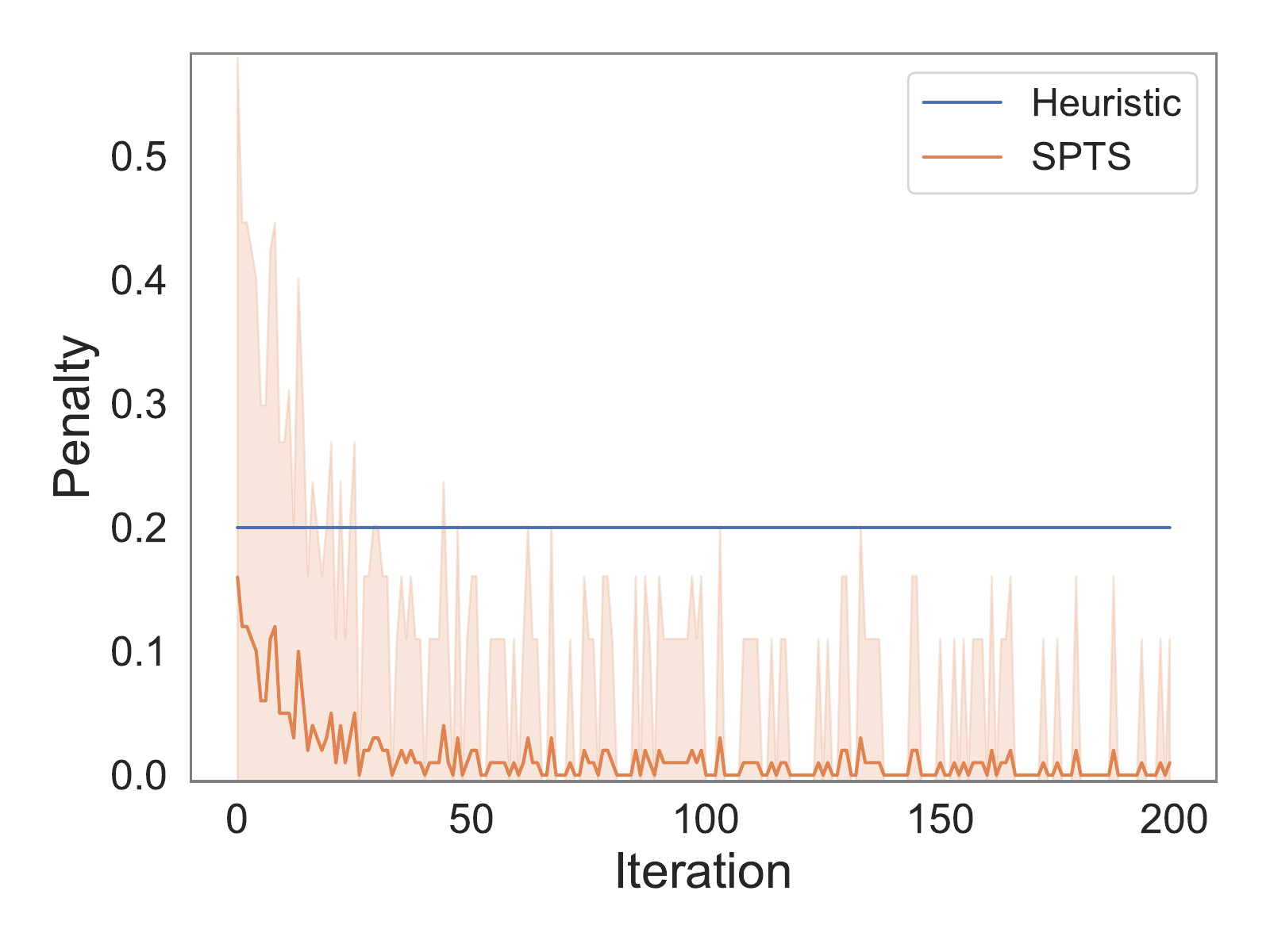}{Penalty}{0}{70}
\sptslearncurve{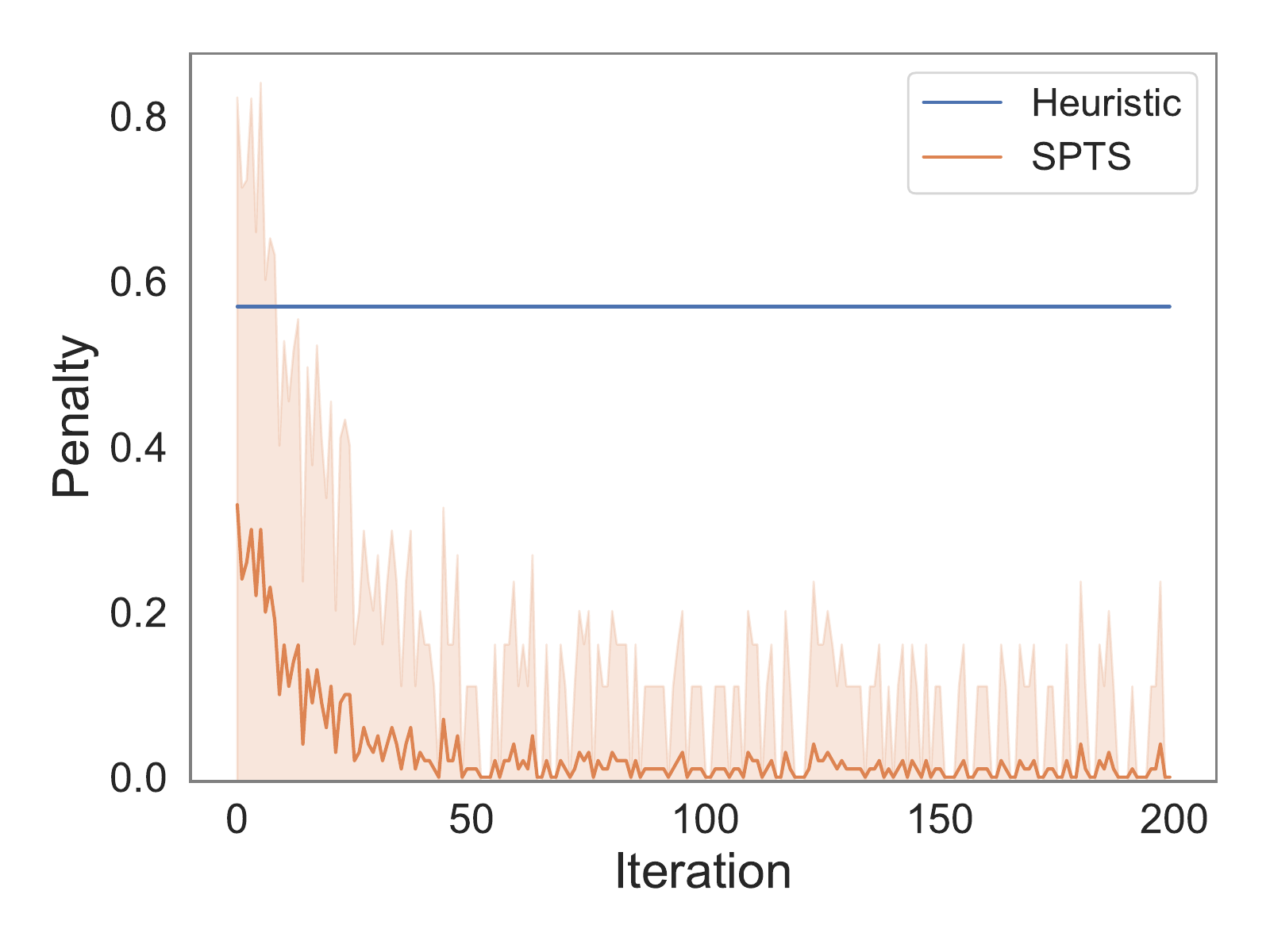}{Penalty}{0}{80}
\sptslearncurve{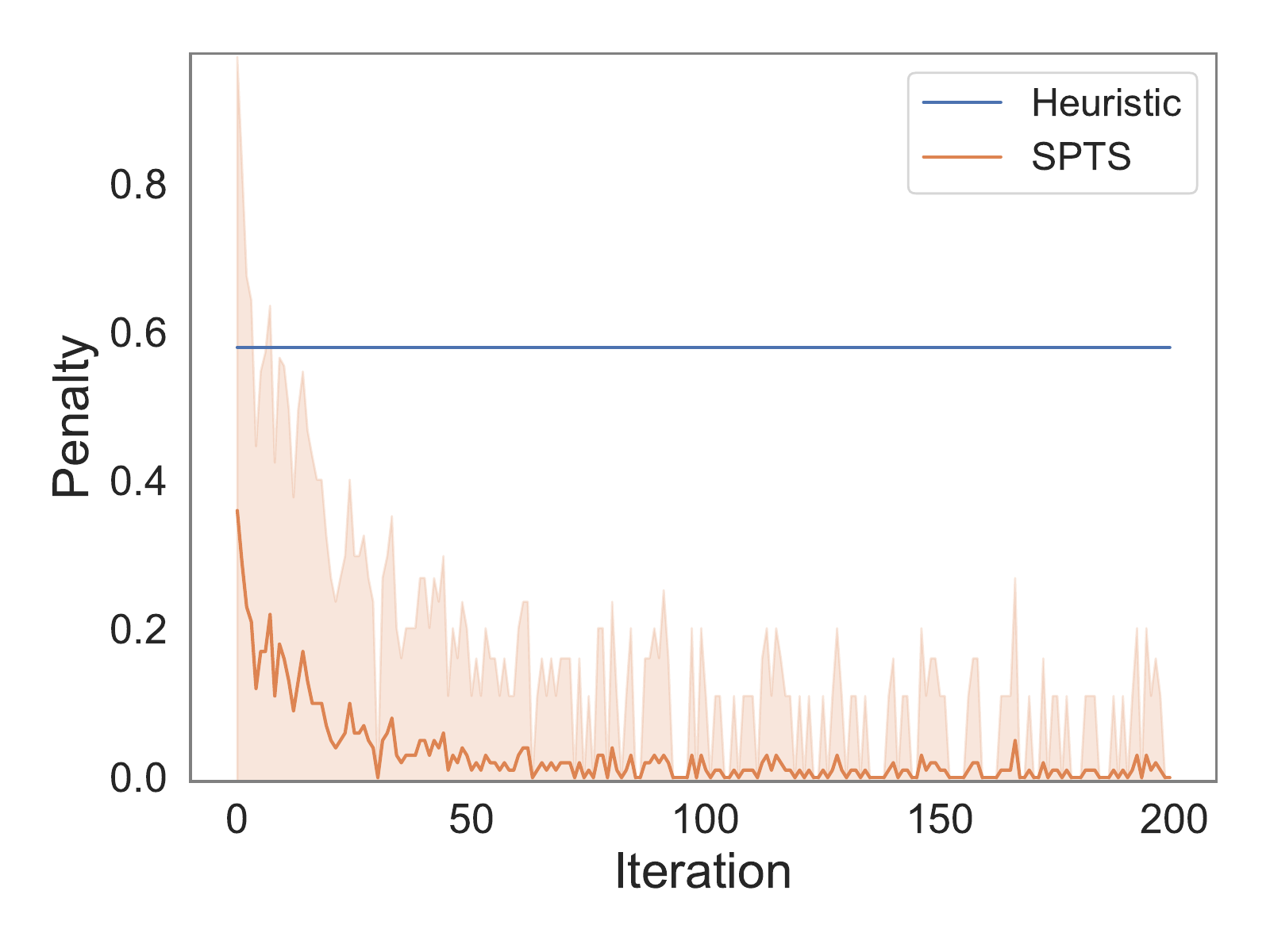}{Penalty}{0}{90}
\sptslearncurve{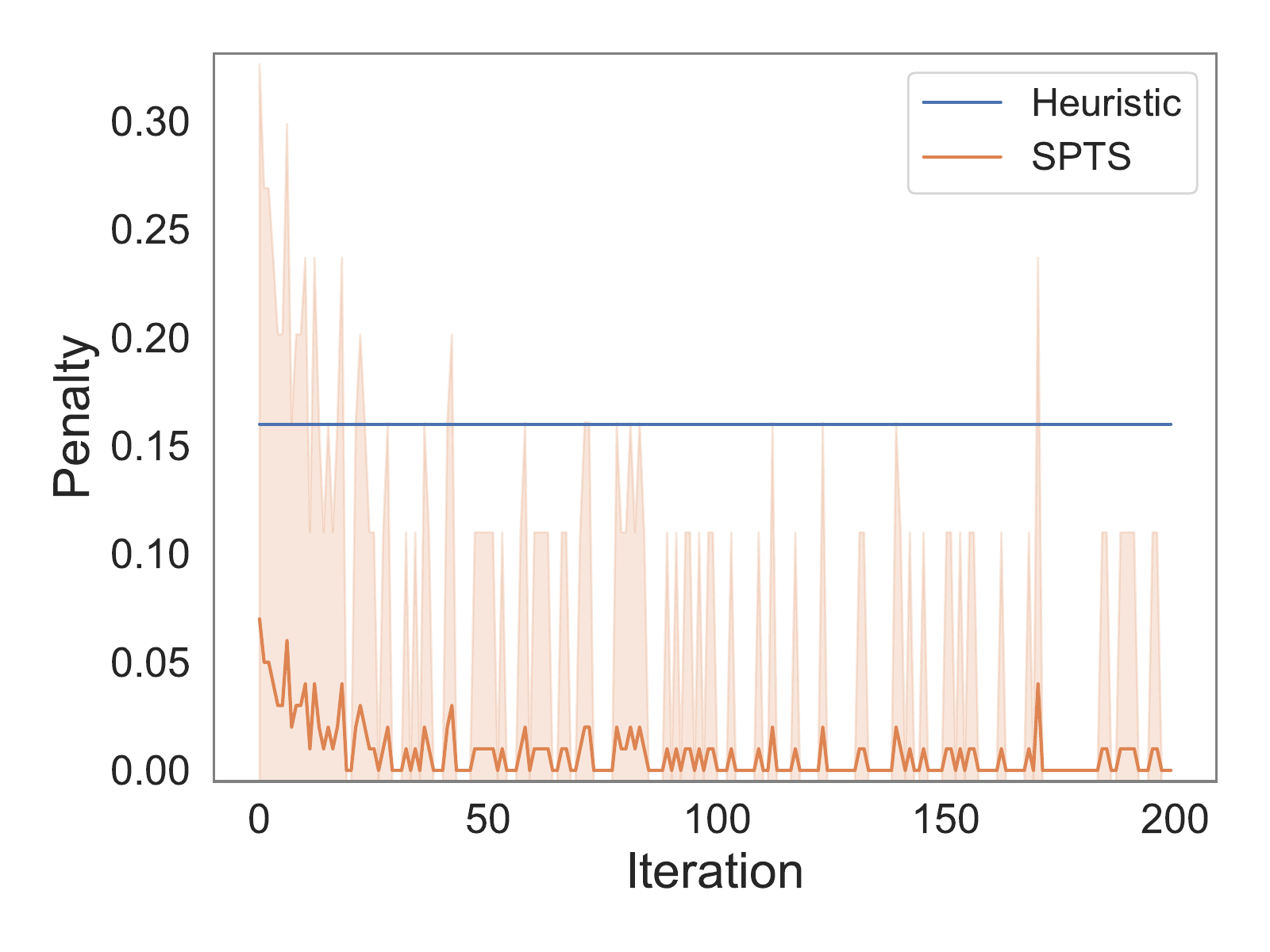}{Penalty}{0}{100}
\sptslearncurve{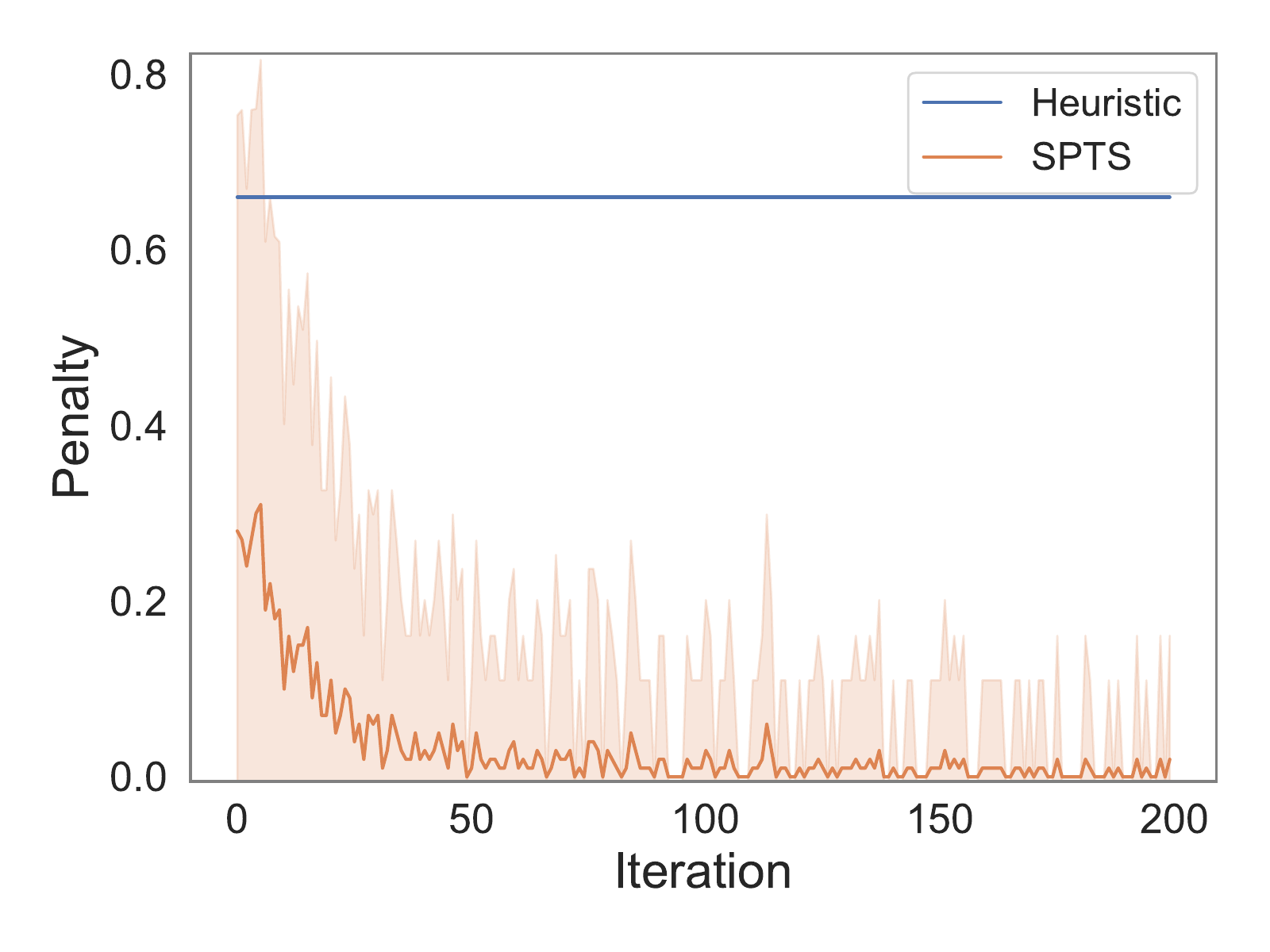}{Penalty}{30}{60}
\sptslearncurve{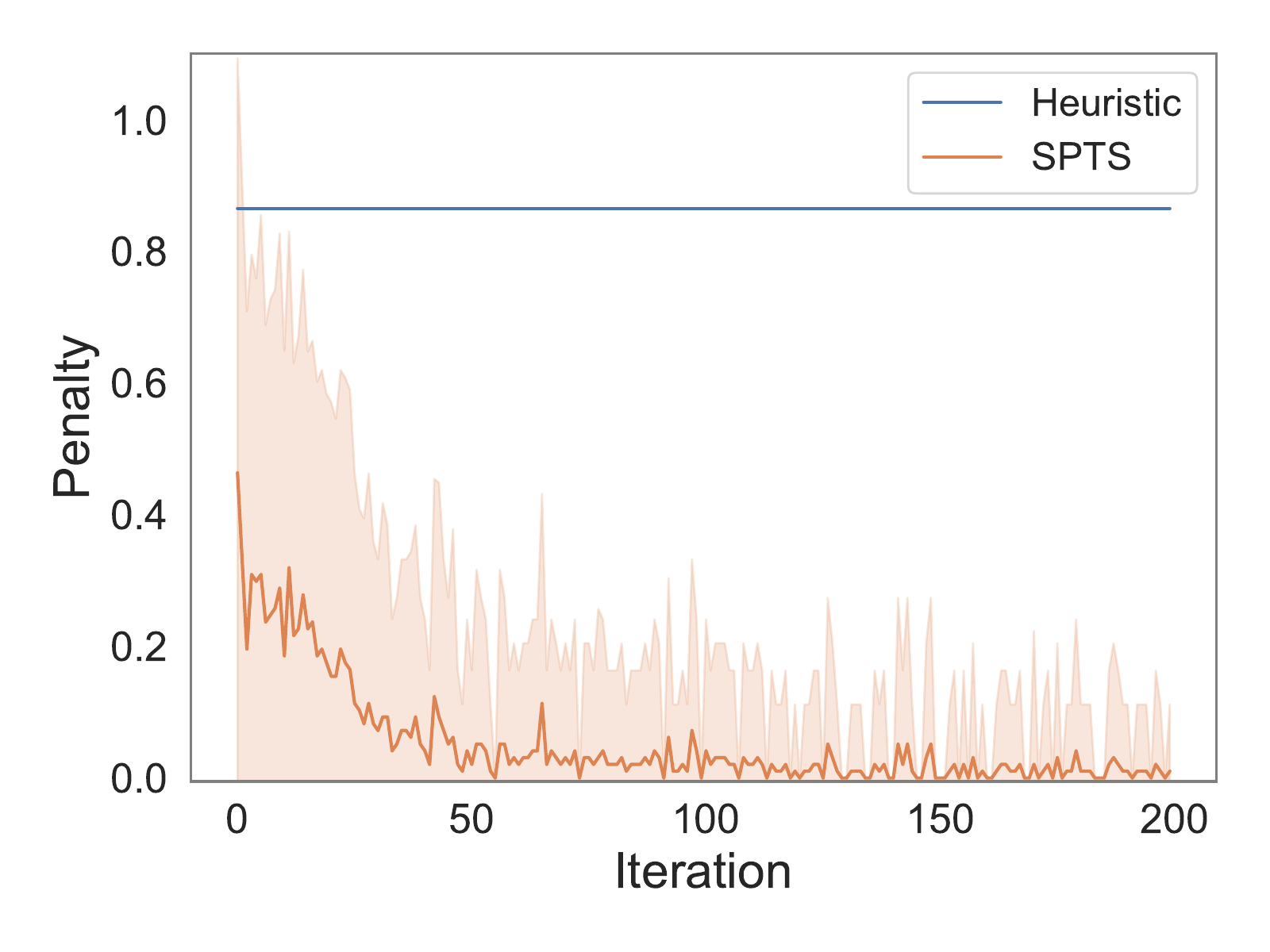}{Penalty}{30}{70}
\sptslearncurve{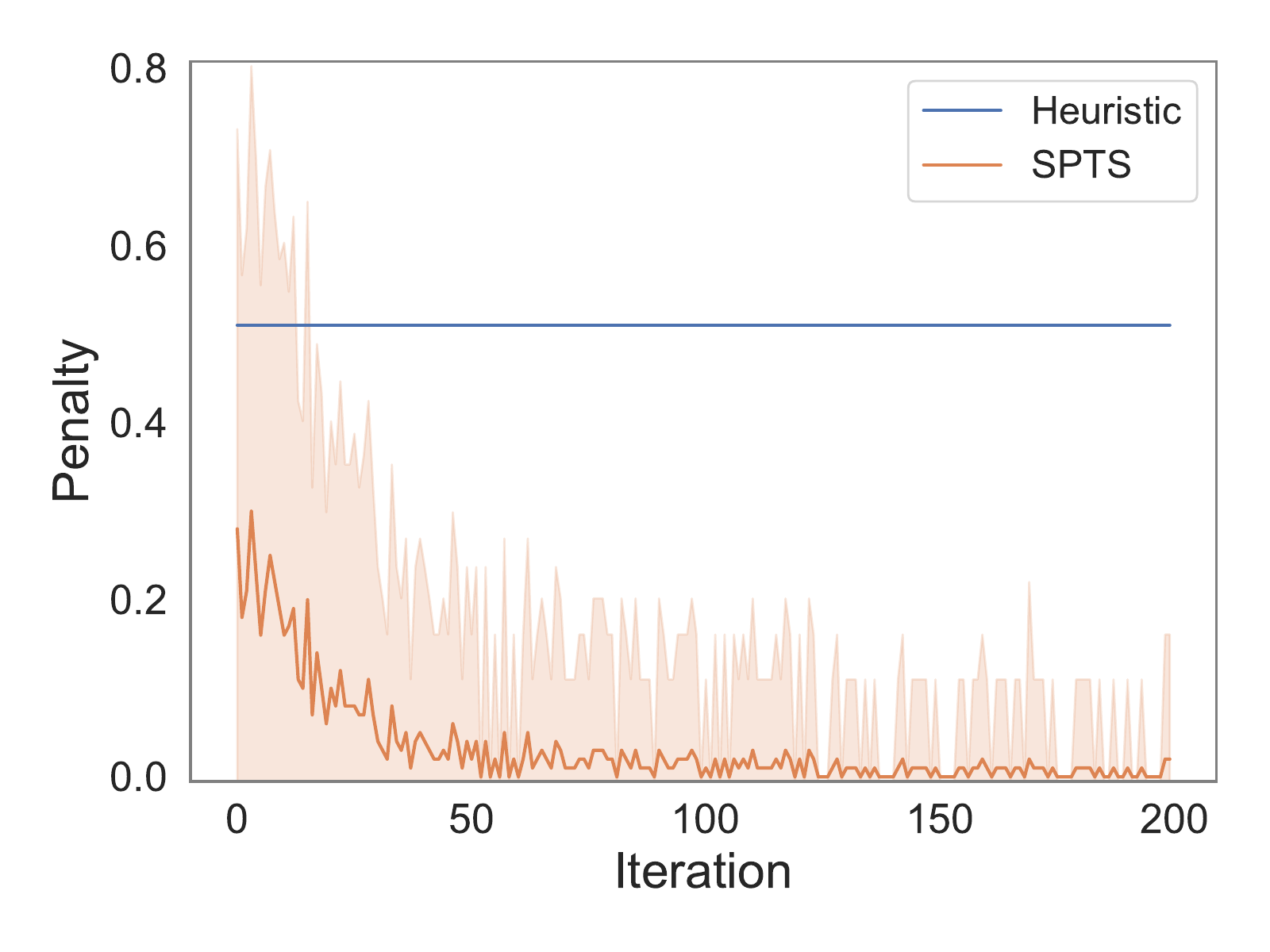}{Penalty}{30}{80}
\sptslearncurve{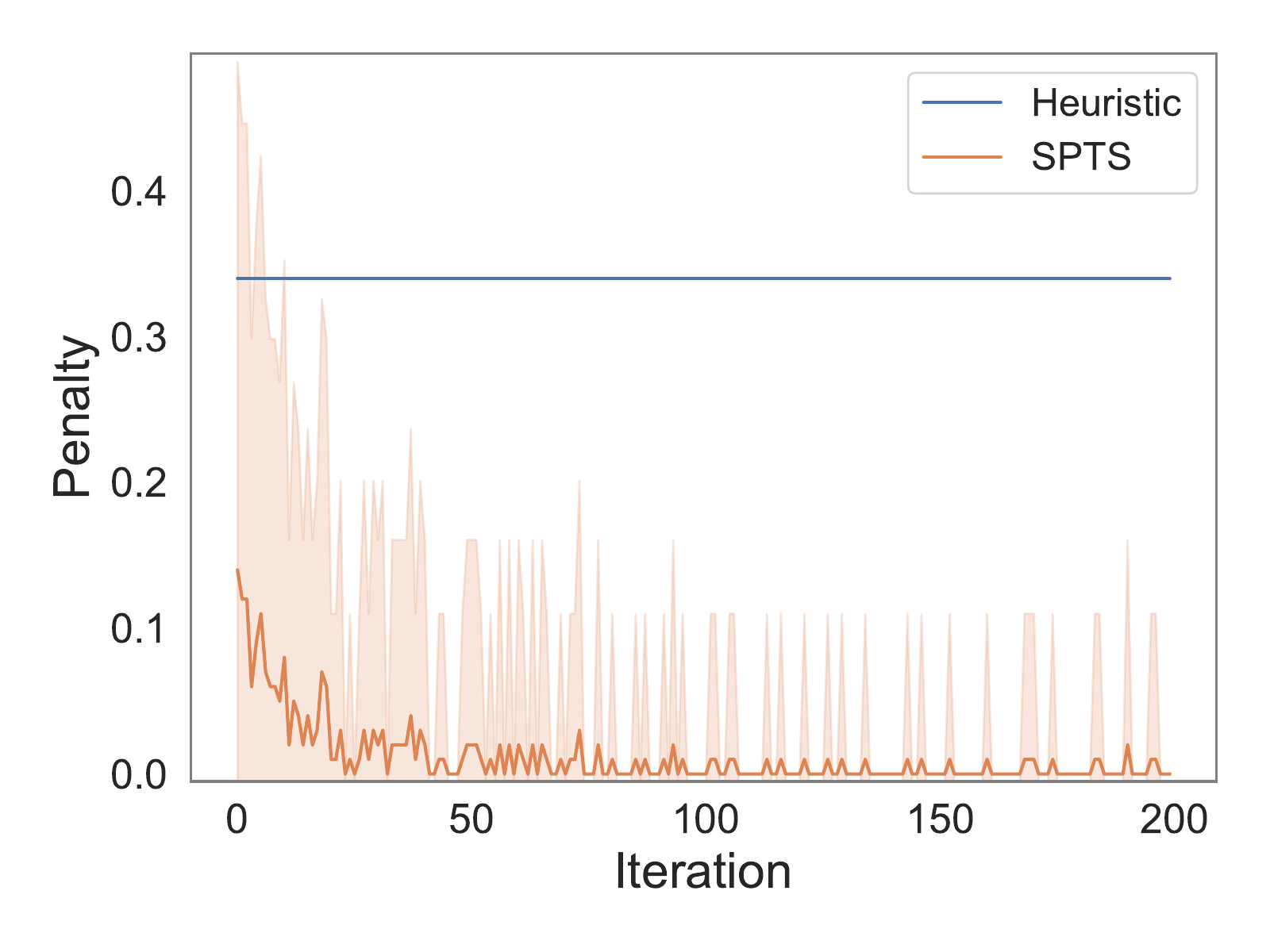}{Penalty}{30}{90}
\sptslearncurve{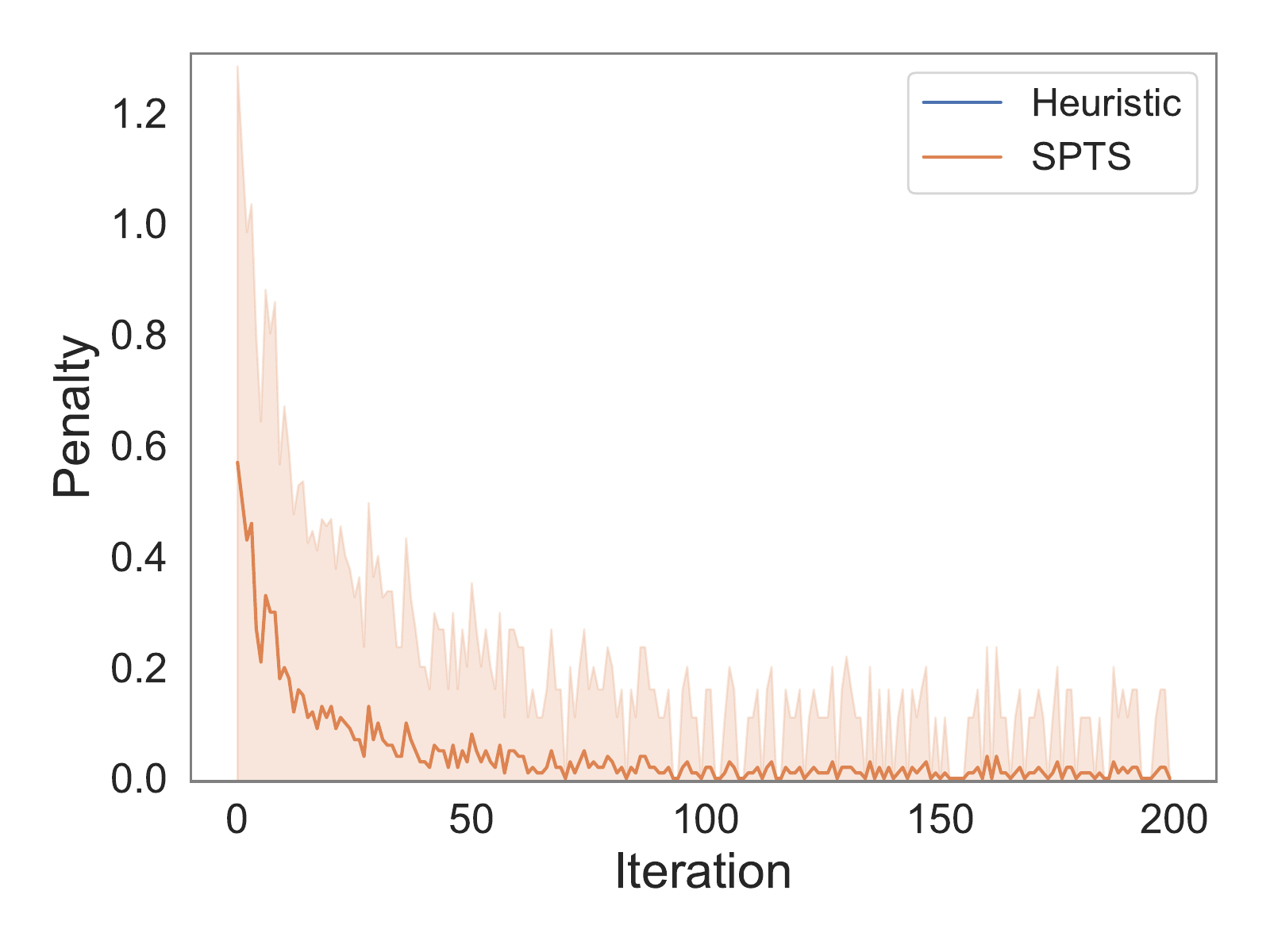}{Penalty}{30}{100}
\clearpage

\section*{Appendix B -- Best Set-Point Configurations}
The best solutions found by SPTS have no penalty. Therefore, we only show the total penalty for the best solutions obtained by the heuristic approach.

\sptsbest{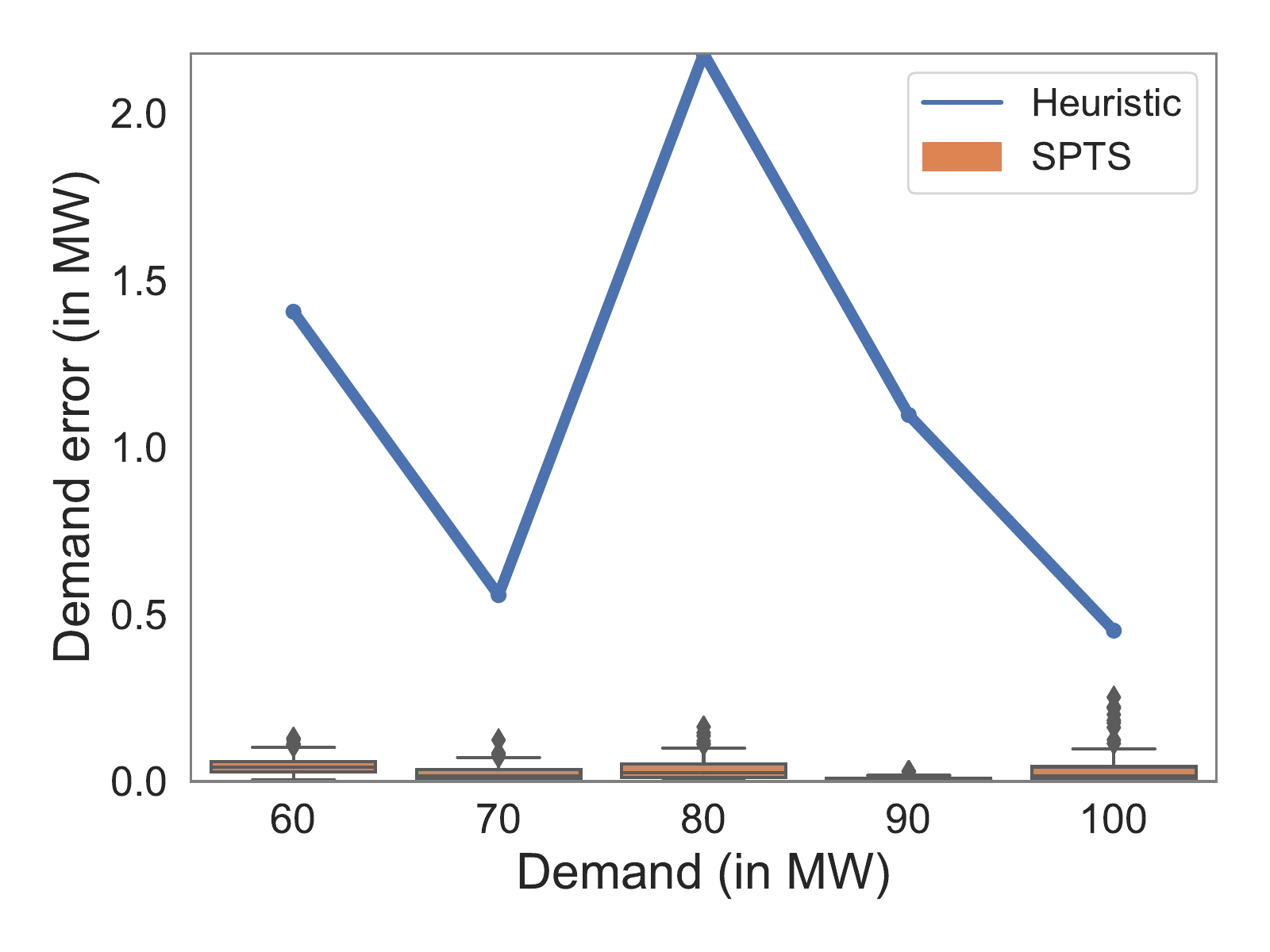}{Demand error of best solution}{0}
\sptsbest{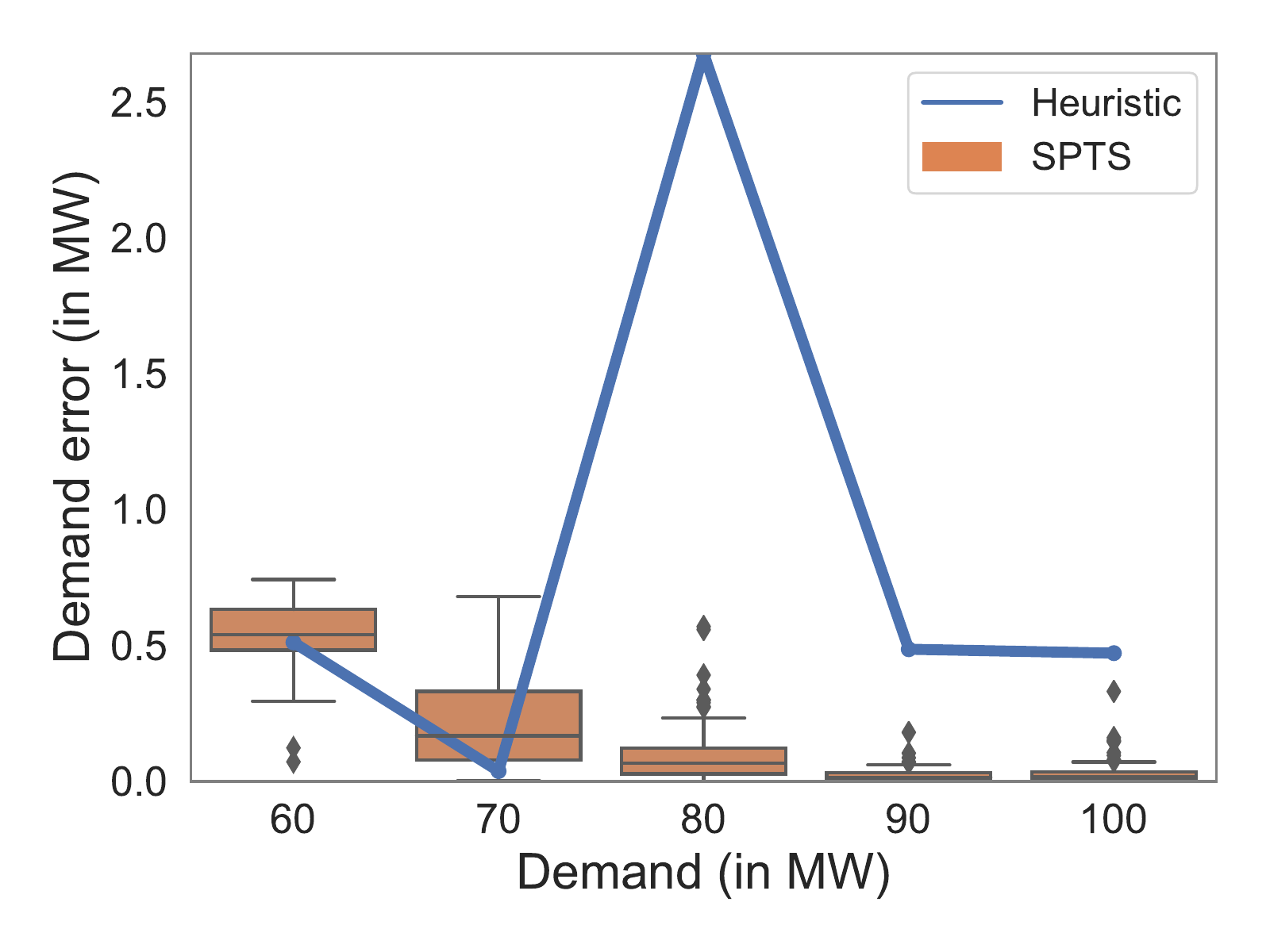}{Demand error of best solution}{30}
\clearpage
\sptsbest{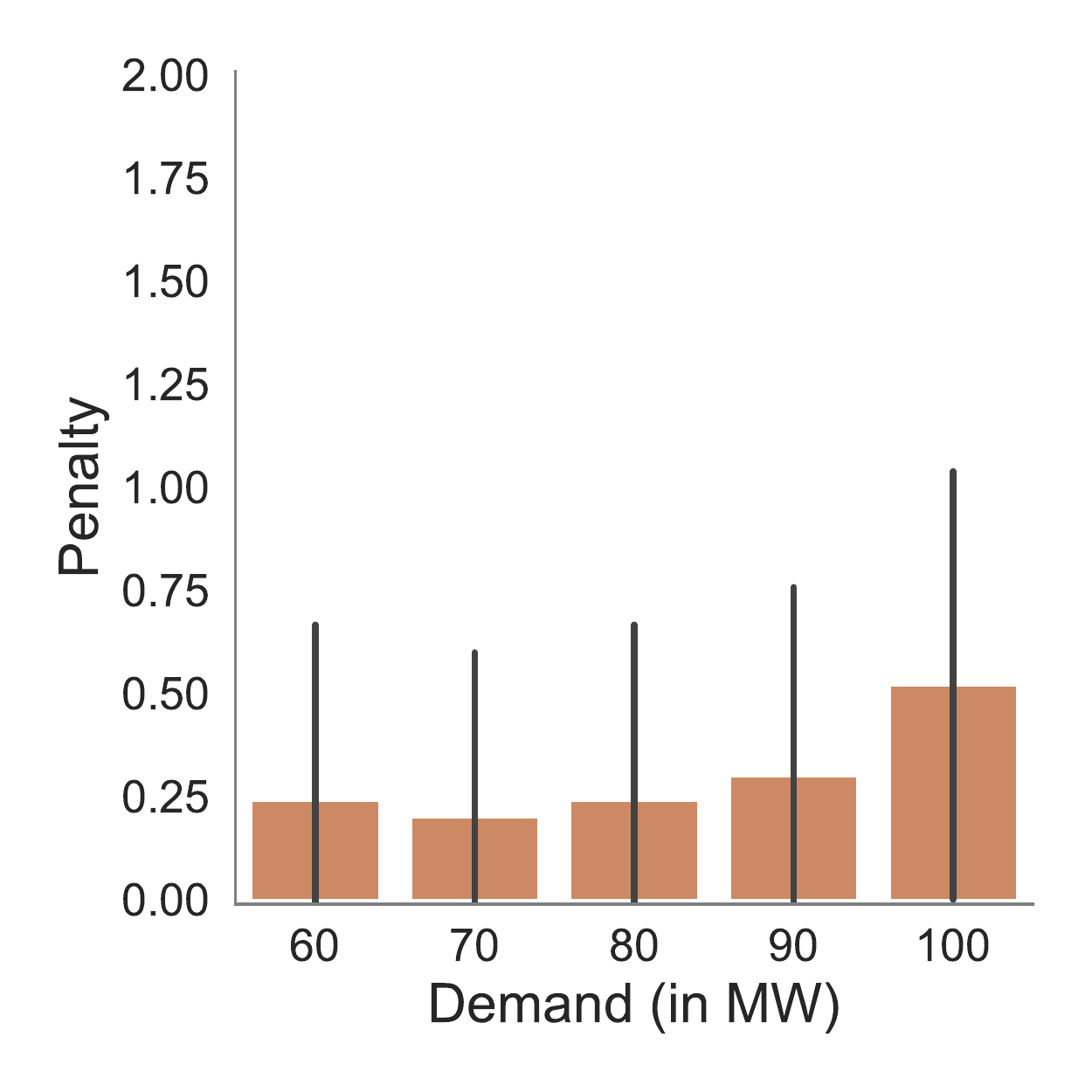}{Penalty of best solution found by the heuristic approach}{0}
\sptsbest{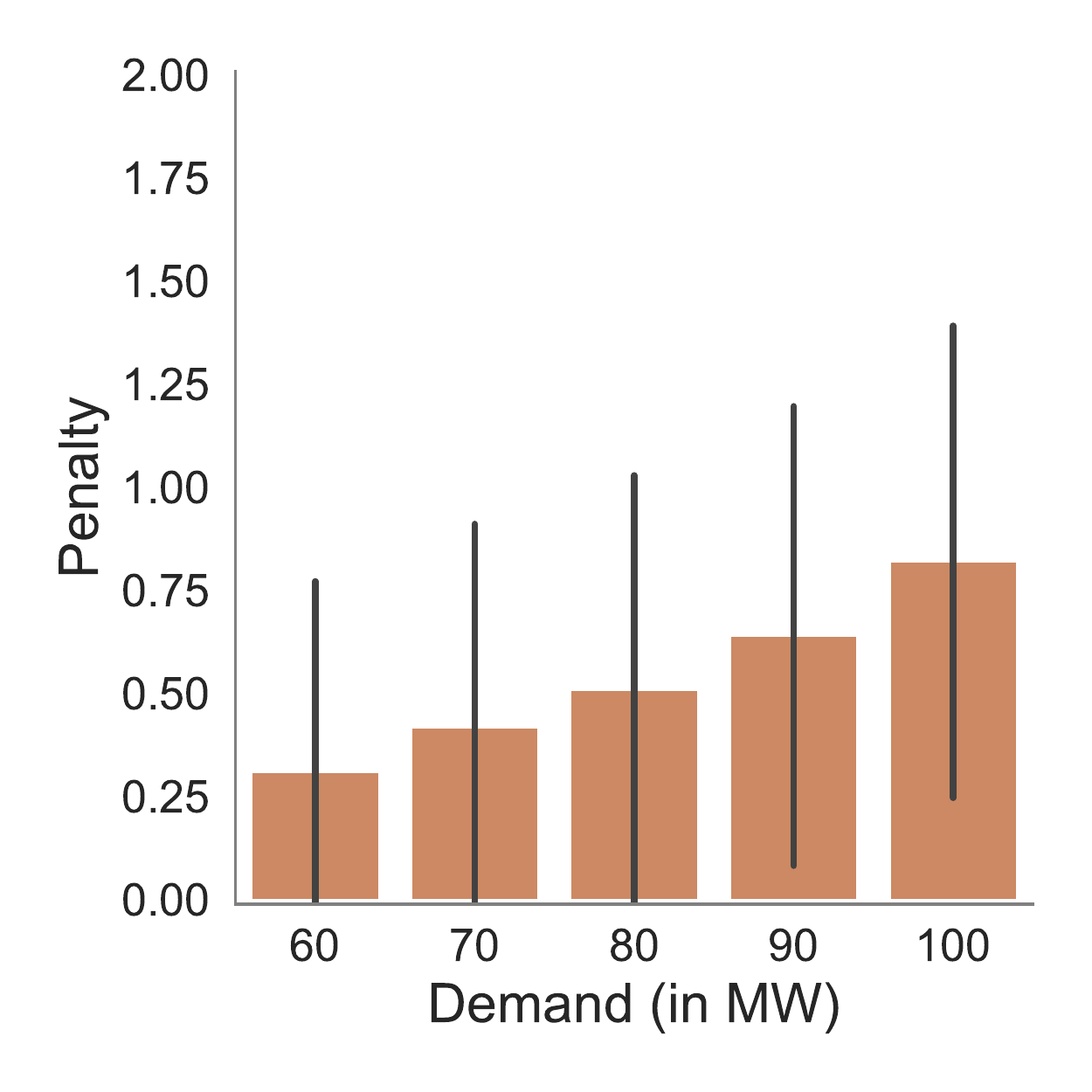}{Penalty of best solution found by the heuristic approach}{30}
\clearpage

\section*{Appendix C -- Farm-Wide Power Productions}

\subsection*{Appendix C.1 -- SPTS}
\sptsfarmpower{farm_power_225_60}{SPTS}{SPTS}{0}{60}
\sptsfarmpower{farm_power_225_70}{SPTS}{SPTS}{0}{70}
\sptsfarmpower{farm_power_225_80}{SPTS}{SPTS}{0}{80}
\sptsfarmpower{farm_power_225_90}{SPTS}{SPTS}{0}{90}
\sptsfarmpower{farm_power_225_100}{SPTS}{SPTS}{0}{100}
\sptsfarmpower{farm_power_255_60}{SPTS}{SPTS}{30}{60}
\sptsfarmpower{farm_power_255_70}{SPTS}{SPTS}{30}{70}
\sptsfarmpower{farm_power_255_80}{SPTS}{SPTS}{30}{80}
\sptsfarmpower{farm_power_255_90}{SPTS}{SPTS}{30}{90}
\sptsfarmpower{farm_power_255_100}{SPTS}{SPTS}{30}{100}
\clearpage

\subsection*{Appendix C.2 -- Heuristic}
\sptsfarmpower{farm_power_225_60}{baseline}{the heuristic approach}{0}{60}
\sptsfarmpower{farm_power_225_70}{baseline}{the heuristic approach}{0}{70}
\sptsfarmpower{farm_power_225_80}{baseline}{the heuristic approach}{0}{80}
\sptsfarmpower{farm_power_225_90}{baseline}{the heuristic approach}{0}{90}
\sptsfarmpower{farm_power_225_100}{baseline}{the heuristic approach}{0}{100}
\sptsfarmpower{farm_power_255_60}{baseline}{the heuristic approach}{30}{60}
\sptsfarmpower{farm_power_255_70}{baseline}{the heuristic approach}{30}{70}
\sptsfarmpower{farm_power_255_80}{baseline}{the heuristic approach}{30}{80}
\sptsfarmpower{farm_power_255_90}{baseline}{the heuristic approach}{30}{90}
\sptsfarmpower{farm_power_255_100}{baseline}{the heuristic approach}{30}{100}

\end{document}